\documentclass[utf8]{article}

\usepackage{arxiv}
\usepackage{graphicx}
\graphicspath{{img/}}
\usepackage{dcolumn}
\usepackage{bm}
\usepackage[hidelinks]{hyperref}
\usepackage{tikz}
\usetikzlibrary{arrows,arrows.meta,positioning}
\usepackage{xifthen}
\usetikzlibrary{calc}
\usetikzlibrary{shapes.geometric}
\usepackage{url,lineno,microtype,subcaption}
\usepackage[onehalfspacing]{setspace}
\usepackage{placeins}
\usepackage{natbib}
\usepackage{amsmath}
\usepackage{tcolorbox}
\usepackage{xcolor}
\definecolor{boxBackground}{HTML}{FFC857}
\newcolumntype{M}{>{\begin{varwidth}{4cm}}l<{\end{varwidth}}}

\usepackage[colorinlistoftodos,prependcaption,textsize=tiny]{todonotes}
\usepackage{xargs}
\newcommandx{\unsure}[2][1=]{\todo[linecolor=red,backgroundcolor=red!25,bordercolor=red,#1]{#2}}
\newcommandx{\change}[2][1=]{\todo[linecolor=blue,backgroundcolor=blue!25,bordercolor=blue,#1]{#2}}
\newcommandx{\done}[2][1=]{\todo[linecolor=blue,backgroundcolor=green!25,bordercolor=blue,#1]{#2}}
\newcommandx{\info}[2][1=]{\todo[linecolor=OliveGreen,backgroundcolor=OliveGreen!25,bordercolor=OliveGreen,#1]{#2}}
\newcommandx{\improvement}[2][1=]{\todo[linecolor=Plum,backgroundcolor=Plum!25,bordercolor=Plum,#1]{#2}}
\newcommandx{\thiswillnotshow}[2][1=]{\todo[disable,#1]{#2}}
\newcommand{\DKL}{D_\mathrm{KL}}
\newcommand{\El}{E_\mathrm{l}}
\newcommand{\Eexc}{E_\mathrm{exc}}
\newcommand{\Einh}{E_\mathrm{inh}}
\newcommand{\psampled}{p}
\newcommand{\ptarget}{p^*}
\newcommand{\tauref}{\tau_\mathrm{ref}}
\newcommand{\wb}{w_\mathrm{b}}
\newcommand{\zdata}{z^\mathrm{data}}
\newcommand{\zrecon}{z^\mathrm{recon}}

\usepackage{cleveref}
\usepackage{siunitx}
\usepackage[T1]{fontenc}
\usepackage{textcomp}
\usepackage{csquotes}
\usepackage{booktabs}

\newcommand*\circled[1]{\tikz[baseline=(char.base),line width=.5pt,font=\sffamily]{
            \node[shape=circle, fill=black!10, draw, minimum size=.35cm, inner sep=0pt] (char) {#1};}}

\newcommand\newreplacement[2]{\def#1/{#2}}

\newcommand{\bs}{\boldsymbol}

\newreplacement{\BSS}{BrainScaleS}
\newreplacement{\BSSS}{BrainScaleS system}
\newreplacement{\BSSWSS}{BrainScaleS wafer-scale system}
\newreplacement{\BSSWM}{BrainScaleS wafer module}

\newreplacement{\itl}{in-the-loop}
\newreplacement{\Itl}{In-the-loop}
\newreplacement{\ITL}{In-The-Loop}
\newreplacement{\itlnodashes}{in the loop}
\newreplacement{\Itlnodashes}{In the loop}
\newreplacement{\ITLnodashes}{In The Loop}

\author{ 
    \hspace{1mm} \textbf{Akos F. Kungl\,$^{1}$, Sebastian Schmitt\,$^{1}$, Johann Kl\"ahn\,$^{1}$, Paul M\"uller\,$^{1}$, Andreas Baumbach\,$^{1}$, Dominik Dold\,$^{1}$,}\\
    \textbf{Alexander Kugele\,$^{1}$, Eric M\"uller\,$^{1}$, Christoph Koke\,$^{1}$, Mitja Kleider\,$^{1}$, Christian Mauch\,$^{1}$, Oliver Breitwieser\,$^{1}$,}\\
    \textbf{Luziwei Leng\,$^{1}$, Nico G\"urtler\,$^{1}$, Maurice G\"uttler\,$^{1}$, Dan Husmann\,$^{1}$, Kai Husmann\,$^{1}$, Andreas Hartel\,$^{1}$,}\\
    \textbf{Vitali Karasenko\,$^{1}$, Andreas Gr\"ubl\,$^{1}$, Johannes Schemmel\,$^{1}$, Karlheinz Meier\,$^{1}$, Mihai A. Petrovici\,$^{1,2}$} \\
    $^{1}$Kirchhoff-Institute for Physics, Heidelberg University, 69120 Heidelberg, Germany \\
    $^{2}$Department of Physiology, University of Bern, 3012 Bern, Switzerland
}

\begin{document}
\onecolumn

\begin{tcolorbox}[colback=boxBackground,colframe=red!75!black]

This article is bound to be published in Frontiers in Neuromorphic Engineering: \\
\url{https://www.frontiersin.org/articles/10.3389/fnins.2019.01201}. \\
Please cite as:\\
Kungl A. F. et al.\@ (2019) Accelerated Physical Emulation of Bayesian Inference in Spiking Neural Networks. \textit{Front. Neurosci. 13:1201. doi: 10.3389/fnins.2019.01201}

\end{tcolorbox}
\vspace{1cm}

\title{Accelerated physical emulation of Bayesian inference in spiking neural networks}
\renewcommand{\headeright}{}
\renewcommand{\undertitle}{}
\date{2019 November 14th}



\maketitle

\begin{abstract}
The massively parallel nature of biological information processing plays an important role for its superiority to human-engineered computing devices.
In particular, it may hold the key to overcoming the von Neumann bottleneck that limits contemporary computer architectures.
Physical-model neuromorphic devices seek to replicate not only this inherent parallelism, but also aspects of its microscopic dynamics in analog circuits emulating neurons and synapses.
However, these machines require network models that are not only adept at solving particular tasks, but that can also cope with the inherent imperfections of analog substrates.
We present a spiking network model that performs Bayesian inference through sampling on the BrainScaleS neuromorphic platform, where we use it for generative and discriminative computations on visual data.
By illustrating its functionality on this platform, we implicitly demonstrate its robustness to various substrate-specific distortive effects, as well as its accelerated capability for computation.
These results showcase the advantages of brain-inspired physical computation and provide important building blocks for large-scale neuromorphic applications.

\end{abstract}



\section{Introduction}

The aggressive pursuit of Moore's law in conventional computing architectures is slowly but surely nearing its end \citep{waldrop2016chips}, with difficult-to-overcome physical effects, such as heat production and quantum uncertainty, representing the main limiting factor.
The so-called von Neumann bottleneck between processing and memory units represents the main cause, as it effectively limits the speed of these largely serial computation devices.
The most promising solutions come in the form of massively parallel devices, many of which are based on brain-inspired computing paradigms \citep{indiveri2011neuromorphic,furber2016large}, each with its own advantages and drawbacks.

Among the various approaches to such neuromorphic computing, one class of devices is dedicated to the physical emulation of cortical circuits: not only do they instantiate neurons and synapses that operate in parallel and independently of each other, but these units are actually represented by distinct circuits that emulate the dynamics of their biological archetypes \citep{mead1990neuromorphic,indiveri2006vlsi,schemmel2010wafer,jo2010nanoscale,pfeil2013six,qiao2015reconfigurable,chang2016demonstration,wunderlich2018demonstrating}.
Some important advantages of this approach lie in their reduced power consumption and enhanced speed compared to conventional simulations of biological neuronal networks, which represent direct payoffs of replacing the resource-intensive numerical calculation of neuro-synaptic dynamics with the physics of the devices themselves.

However, such computation with analog dynamics, without the convenience of binarization, as used in digital devices, has a downside of its own: variability in the manufacturing process (fixed pattern noise) and temporal noise both lead to reduced controllability of the circuit dynamics.
Additionally, one relinquishes much of the freedom permitted by conventional algorithms and simulations, as one is confined by the dynamics and parameter ranges cast into the silicon substrate.
The main challenge of exploiting these systems therefore lies in designing performant network models using the available components while maintaining a degree of robustness towards the substrate-induced distortions.
Just like for the devices themselves, inspiration for such models often comes from neuroscience, as the brain needs to meet similar demands.

With accumulating experimental evidence \citep{berkes2011spontaneous,pouget2013probabilistic,orban2016neural,haefner2016perceptual}, the view of the brain itself as an analytical computation device has shifted.
The stochastic nature of neural activity in vivo is being increasingly regarded as an explicit computational resource rather than a nuisance that needs to be dealt with by sophisticated error-correcting mechanisms or by averaging over populations.
Under the assumption that stochastic brain dynamics reflect an ongoing process of Bayesian inference in continuous time, the output variability of single neurons can be interpreted as a representation of uncertainty.
Theories of neural sampling \citep{buesing2011neural,hennequin2014fast,aitchison2016hamiltonian,petrovici2016stochastic,kutschireiter2017nonlinear} provide an analytical framework for embedding this type of computation in spiking neural networks.

In this paper we describe the realization of neural sampling with networks of leaky integrate-and-fire neurons \citep{petrovici2016stochastic} on the BrainScaleS accelerated neuromorphic platform \citep{schemmel2010wafer}.
With appropriate training, the variability of the analog components can be naturally compensated and incorporated into a functional network structure, while the network's ongoing dynamics make explicit use of the analog substrate's intrinsic acceleration for Bayesian inference (\cref{sec:experimentalSetup}).
We demonstrate sampling from low-dimensional target probability distributions with randomly chosen parameters (\cref{sec:distr}) as well as inference in high-dimensional spaces constrained by real-world data, by solving associated classification and constraint satisfaction problems (pattern completion, \cref{sec:datasets}).
All network components are fully contained on the neuromorphic substrate, with external inputs only used for sensory evidence (visual data).
Our work thereby contributes to the search for novel paradigms of information processing that can directly benefit from the features of neuro-inspired physical model systems.

\section{Methods}
\subsection{The BrainScaleS system}
\label{sec:bss}

\begin{figure}
  \centering
  \includegraphics[width=\textwidth]{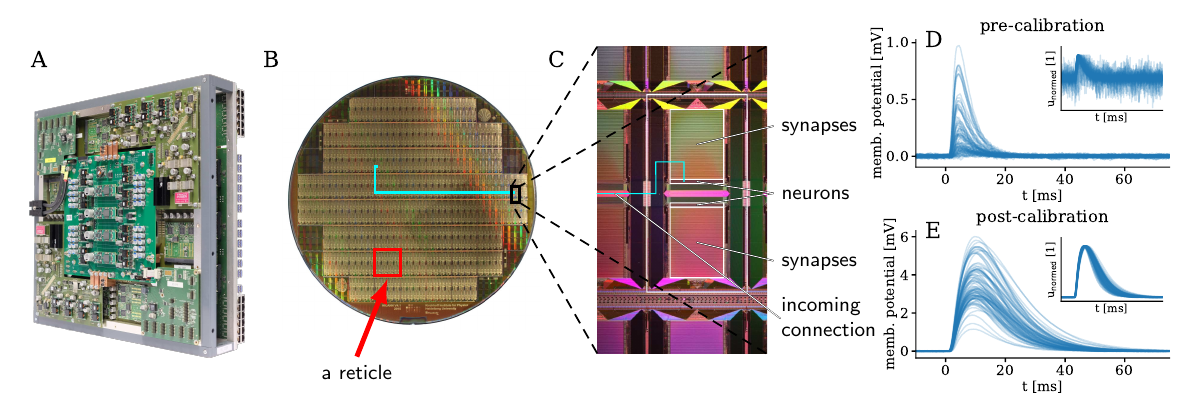}
  \hspace{\fill}
  \caption{%
    \protect{\textbf{(A)}} Photograph of a fully assembled wafer module of the BrainScaleS system (dimensions: \SI{50}{\centi\meter} \texttimes{} \SI{50}{\centi\meter} \texttimes{} \SI{15}{\centi\meter}).
    One module hosts 384 HICANN chips on 48 reticles, with 512 physical neurons per chip and 220 synapse circuits per neuron.
    The wafer itself lies at the center of the module and is itself not visible.
    \num{48} FPGAs are responsible for I/O and experiment control.
    Support PCBs provide power supply for the on-wafer circuits as well as access to neuron membrane voltages.
    The connectors for inter-wafer (sockets resembling USB-A) and off-wafer/host connectivity (Gigabit-Ethernet sockets) are distributed over all four edges of the main PCB.
    Mechanical stability is provided by an aluminum frame.
    \protect{\textbf{(B)}} The wafer itself is composed of 48 reticles (e.g., red rectangle), each containing 8 HICANN chips (e.g., black rectangle, enlarged in C).
    Inter-reticle connectivity is added in a post-processing step.
    \textbf{(C)} On a single HICANN chip, the largest area is occupied by the two synapse matrices which instantiate connections to the neurons positioned in the neuron array.
    {
    \textbf{(D-E)} Postsynaptic potentials (PSPs) measured on 100 different neuron membranes using the same parameter settings before \textbf{(D)} and after \textbf{(E)} calibration.
    The insets show the height-normalized PSPs.
    The calibration serves two purposes.
    First, it provides a translation rule between the desired neuron parameters and the technical parameters set on the hardware.
    In this case, it brings the time constants $\tau_\mathrm{mem}$ and $\tau_\mathrm{syn}$ close to the target of \SI{8}{\milli\second}, as evidenced by the small spread of the normalized PSPs.
    Second, in the absence of such a translation rule, it sets the circuits to their correct working points.
    Here, this happens for the synaptic weights: after calibration, PSP heights are, on average closer to the target working point of \SI{3}{\milli\volt}, but they remain highly diverse due to the variability of the substrate.
    For more details see \cite{schmitt2017neuromorphic}.
    }
    The PSPs are averaged over 375 presynaptic spikes and smoothed with a Savitzky-Golay filter \citep{savitzky1964smoothing} to eliminate readout noise.
    The time-constants are given in the biological domain, but they are $10^4$ faster on the system.
  }
  \label{fig:wafer_module}
\end{figure}

BrainScaleS \citep{schemmel2010wafer} is a mixed-signal neuromorphic system, realized in \SI{180}{nm} CMOS technology, that emulates networks of spiking neurons.
Each BrainScaleS wafer module consists of a \SI{20}{\centi\meter} silicon wafer with 384 HICANN (High Input Count Analog Neural Network) chips, see \cref{fig:wafer_module}~A.
On each chip, 512 analog circuits emulate the adaptive exponential integrate-and-fire (AdEx) model \citep{brette2005adaptive,millner2010AdEx} of spiking neurons with conductance-based synapses.
The dynamics evolve with an acceleration factor of $10^4$ with respect to biological time, i.e., all specific time constants (synaptic, membrane, adaptation) are approximately $10^4$ times smaller than typical corresponding values found in biology \citep{schemmel2010wafer,petrovici2014characterization}.
To preserve compatibility with related literature \citep{petrovici2016stochastic,schmitt2017neuromorphic,leng2018spiking,dold2018stochasticity}, { we refer to system parameters in the biological domain unless specified otherwise, e.g., a membrane time constant given as \SI{10}{ms} is actually accelerated to \SI{1}{\micro\second} on the chip.}

The parameters of the neuron circuits are stored in analog memory cells (floating gates) with \SI{10}{bit} resolution, and the synaptic weights are stored in \SI{4}{bit} SRAM \citep{schemmel2010wafer}.
{ The analog memory cells are similar to the ones in \cite{lande1996analog}, and they are described in \cite{loock2006evaluierung} and \cite{millner2012development}.}

Spike events are transported digitally and can reach all other neurons on the wafer with the help of an additional redistribution layer that instantiates an on-wafer circuit-switched network \citep{zoschkeguettler2017rdlembedding} (\cref{fig:wafer_module}~B).

Because of mismatch effects (fixed-pattern noise) inherent to the substrate, the response to incoming stimuli varies from neuron to neuron (\cref{fig:wafer_module}~D).
In order to bring all neurons into the desired regime and reduce the neuron-to-neuron response variability, we employ a standard calibration procedure that is performed only once, during the commissioning of the system \citep{schmitt2017neuromorphic,petrovici2017robustness}.
Nevertheless, even after calibration, a significant degree of diversity persists (\cref{fig:wafer_module}~E).
The emulation of functional networks that do not rely on population averaging therefore requires appropriate training algorithms (\cref{sec:datasets}).

\subsection{Sampling with leaky integrate-and-fire neurons}

\begin{figure}
  \centering
    \includegraphics[width=0.65\textwidth]{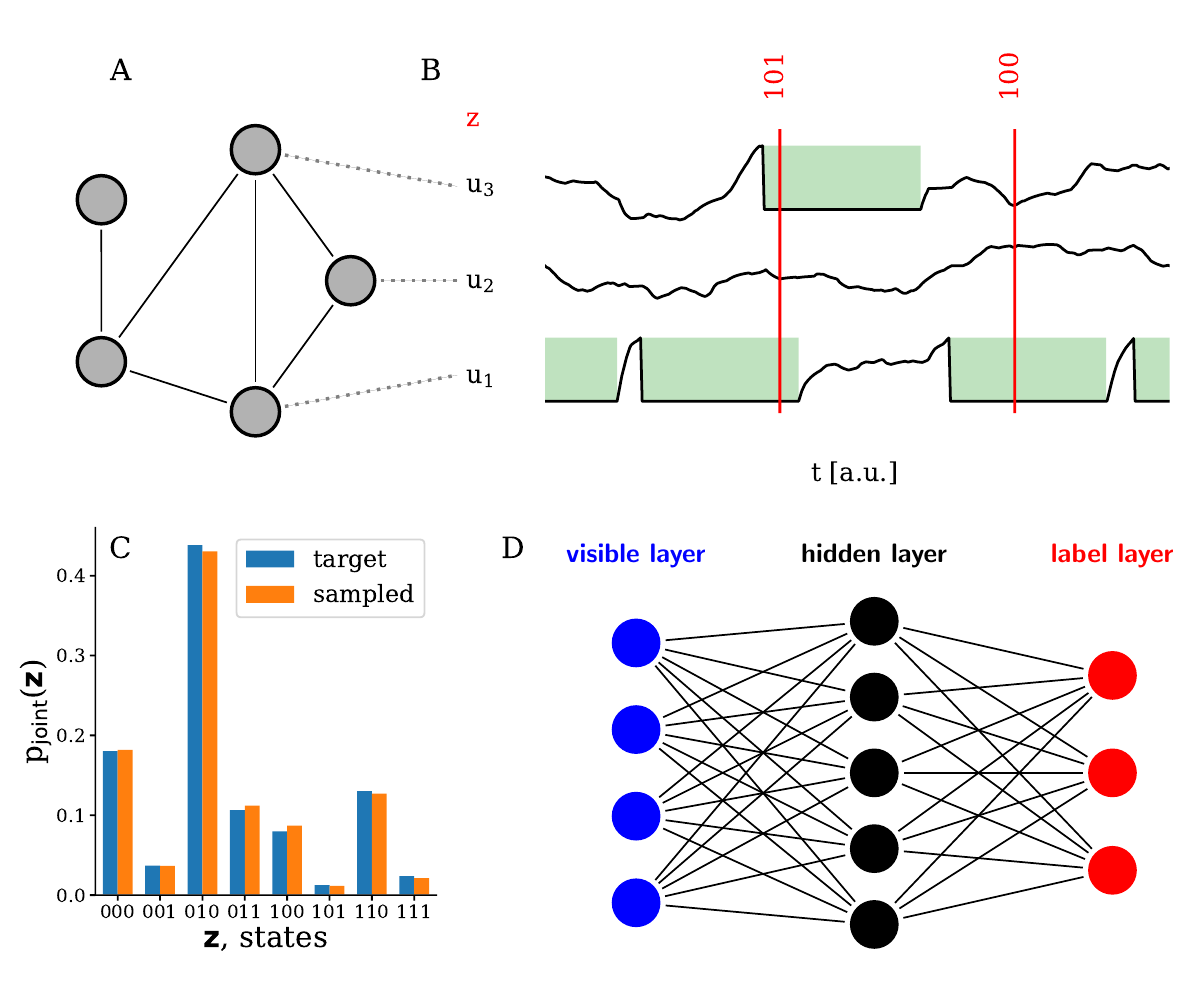}
  \caption{
            \textbf{Sampling with leaky integrate-and-fire (LIF) neurons}.
            \textbf{(A)} Schematic of a spiking sampling network (SSN) with 5 neurons.
            Each line represents two reciprocal synaptic connections with equal weights.
            \textbf{(B)} Example membrane potentials of three neurons in the network.
            Following a spike, the refractory mechanism effectively clamps the membrane potential to the reset value for a duration $\tauref$.
            During this time, the RV corresponding to that neuron is in the state $z=1$ {(marked in green)}.
            At any point in time, the state sampled by the network can therefore be constructed directly from its output spikes and the refractory time $\tau_\mathrm{ref}$ of the neurons.
            {
            \textbf{(C)} Probability distribution sampled by an SSN with three neurons as compared to the target distribution.
            }
            \textbf{(D)} Based on this framework \citep{petrovici2016stochastic}, hierarchical sampling networks can be built, which can be trained on real-world data.
            {Each line represents a reciprocal connection (two synapses) between the connected neurons.}
            }
  \label{fig:liftheory}
\end{figure}

The theory of sampling with leaky integrate-and-fire neurons \citep{petrovici2016stochastic} describes a mapping between the dynamics of a population of neurons with conductance-based synapses (equations given in \cref{table:network}) and a Markov-chain Monte Carlo sampling process from an underlying probability distribution over binary random variables (RVs).
Each neuron in such a sampling network corresponds to one of these RVs: if the $k$-th neuron has spiked in the recent past and is currently refractory, then it is considered to be in the \emph{on-state} $z_k=1$, otherwise it is in the \emph{off-state} $z_k=0$ (\cref{fig:liftheory}~A,~B).
With appropriate synaptic parameters, such a network can approximately sample from a Boltzmann distribution defined by
\begin{align}
p^*(\bs z) &= \frac{1}{Z} \exp \left ( \frac{1}{2} \bs z^T \bs W \bs z + \bs z^T \bs b  \right) \; ,
\end{align}
where $Z$ is the partition sum, $\bs W$ a symmetric, zero-diagonal effective weight matrix and $b_i$ the effective bias of the $i$-th neuron.

In the original model, each neuron receives excitatory and inhibitory Poisson input.
{
This plays two important roles: it transforms a deterministic LIF neuron into a stochastic firing unit and induces a high-conductance state, with an effective membrane time constant that is much smaller than other time constants in the system: $\tau_\mathrm{eff} \gg \tau_\mathrm{syn}, \tau_\mathrm{ref}$ - \citep[see, e.g., ][]{destexhe2003high,petrovici2016form}, which symmetrizes the neural activation function, as explained in the following.
The activation function of an LIF neuron without noise features a sharp onset, but only a slow converge to its maximum value, hence being highly asymmetric around the point of \SI{50}{\%} activity.
Background Poisson noise smears out the onset of the activation function, while the reduced membrane time constant accelerates the convergence to the maximum, making the activation function more symmetric and thus more similar to a logistic function, which is a prerequisite for this form of sampling.
For the explicit derivation see \cite{petrovici2016stochastic} and \cite{petrovici2016form}.
}
A mapping of this activation function to the abovementioned logistic function $1/[1 + \exp(-x)]$ provides the translation from the dimensionless weights and biases of the target distribution to the corresponding biological parameters of the spiking network \citep{petrovici2016form}.

Although different in their dynamics, such sampling spiking networks (SSNs) function similarly to (deep) Boltzmann machines \citep{hinton1984boltzmann}, which makes them applicable to the same class of machine learning problems \citep{leng2018spiking}.
Training can be done using an approximation of wake-sleep algorithm \citep{hinton1995wake,hinton2012practical}, which implements maximum-likelihood learning on the training set:
\begin{align}
\label{eq:wake_sleep}
\Delta b_i &= \eta(\langle z_i \rangle^* - \langle z_i \rangle) \; , \\
\Delta W_{ij} &= \eta (\langle z_i z_j \rangle^* - \langle z_i z_j\rangle) \; ,
\end{align}
where $\langle \cdot \rangle$ and $\langle \cdot \rangle^*$ represent averages over the sampled (model or sleep phase) and target (data or wake phase) distribution, respectively, and $\eta$ is the learning rate.

In order to enable a fully-contained neuromorphic emulation on the BrainScaleS system, the original model had to be modified.
The changes in the network structure, noise generation mechanism and learning algorithm are described in \cref{sec:experimentalSetup}.

For low-dimensional, fully specified target distributions, we used the Kullback-Leibler divergence \citep[DKL,][]{kullback1951information} as a measure of discrepancy between the sampled ($\psampled$) and the target ($\ptarget$) distributions:
\begin{equation}
    \DKL(\psampled \parallel \ptarget) = - \sum_{z_i \in \Omega} \psampled(z_i) \ln \left (  \frac{\psampled(z_i)}{\ptarget(z_i)} \right )
\end{equation}
This was done in part to preserve comparability with previous studies~\citep{buesing2011neural,petrovici2015sampling,petrovici2016stochastic}, but also because the DKL is the natural loss function for maximum likelihood learning.
For visual datasets, we used the error rate (ratio of misclassified images in the test set) for discriminative tasks and the mean squared error (MSE) between reconstruction and original image for pattern completion tasks.
The MSE is defined as
\begin{equation}
\text{MSE} = \frac{1}{N_\text{pixels}} \sum_{k=1}^{N_\text{pixels}}\left ( \zdata_k - \zrecon_k \right)^2 \; ,
\end{equation}
where $\zdata_k$ is the reference data value, $\zrecon_k$ is the model reconstruction and the sum goes over the $N_\text{pixels}$ pixels to be reconstructed by the SSN.

\subsection{Experimental setup}
\label{sec:experimentalSetup}

The physical emulation of a network model on an analog neuromorphic substrate is not as straightforward as a software simulation, as it needs to comply with the constraints imposed by the emulating device.
Often, it may be tempting to fine-tune the hardware to a specific configuration that fits one particular network, e.g., by selecting specific neuron and synapse circuits that operate optimally given a particular set of network parameters, or by manually tweaking individual hardware parameters after the network has been mapped and trained on the substrate.
Here, we explicitly refrained from any such interventions in order to guarantee the robustness and scalability of our results.

All experiments were carried out on a single module of the BrainScaleS system using a subset of the available HICANN chips.
The network setup was specified in the BrainScaleS-specific implementation of PyNN \citep{davison2009pynn} and the standard calibration \citep{schmitt2017neuromorphic} was used to set the analog parameters.
The full setup consisted of two main parts: the SSN and the source of stochasticity.

\begin{figure}
  \centering
  \includegraphics[width=\textwidth]{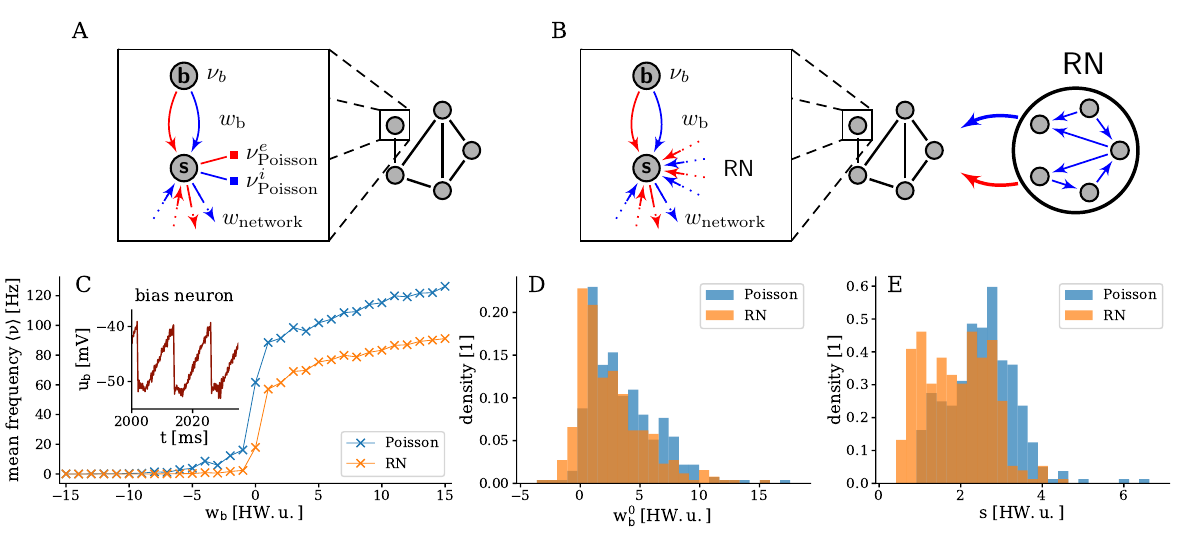}

  \caption{\textbf{Experimental setup.}
    Each sampling unit is instantiated by a pair of neurons on the hardware.
    The bias neuron \protect\circled{\textbf{b}} is configured with a suprathreshold leak potential and generates a regular spike train that impinges on the sampling neuron \protect\circled{\textbf{s}}, thereby serving as a bias, controlled by $\wb$.
    \textbf{(A)} As a benchmark, we provided each sampling neuron with private, off-substrate Poisson spike sources.
    \textbf{(B)} Alternatively, in order to reduce the I/O load, the noise was generated by a random network (RN).
    The RN consisted of randomly connected inhibitory neurons with $E_\mathrm{leak}>V_\mathrm{thresh}$.
    Connections were randomly assigned, such that each sampling neuron received a fixed number of excitatory and inhibitory presynaptic partners (\cref{table:network}).
    \textbf{(C)} Exemplary activation function (mean firing frequency) of a single sampling neuron with Poisson noise and with an RN as a function of the bias weight.
    The standard deviation of the the trial-to-trial variability is on the order of \SI{0.1}{Hz} for both activation functions, hence the error bars are to small to be shown.
    The inset shows the membrane trace of the corresponding bias neuron.
    \textbf{(D-E)} {The figures show histograms over all neurons in a sampling network on a calibrated BrainScaleS system.}
    The width $s$ and the midpoint $w^0_\mathrm{b}$ of the activation functions with Poisson noise and with an RN are calculated by fitting the logistic function $\langle \nu \rangle = \nu_0/\{1+\exp[-(\wb-w^0_\mathrm{b})/s]\}$ to the data.
    }
  \label{fig:networkSetup}
\end{figure}

In the original sampling model \citep{petrovici2016stochastic}, in order to affect biases, the wake-sleep algorithm (\cref{eq:wake_sleep}) requires access to at least one reversal potential ($\El$, $\Eexc$, or $\Einh$), which are all controlled by analog memory cells.
Given that rewriting analog memory cells is both less precise and slower than rewriting the SRAM cells controlling the synaptic weights, we modified our SSNs to implement biases by means of synaptic weights.
To this end, we replaced individual sampling neurons by sampling units, each realized using two hardware neurons (\cref{fig:networkSetup} A, B).
Like in the original model, a sampling neuron was set up to encode the corresponding binary RV.
Each sampling neuron was accompanied by a bias neuron set up with a suprathreshold leak potential that ensured regular firing (\cref{fig:networkSetup} C, inset).
Each bias neuron projected to its target sampling neuron with both an excitatory and an inhibitory synapse (with independent weights), thus inducing a controllable offset of the sampling neuron's average membrane potential.
Because excitatory and inhibitory inputs are routed through different circuits for each neuron, two types of synapses were required to allow the sign of the effective bias to change during training.
For larger networks, in order to optimize the allocation of hardware resources, we shared the use of bias neurons among multiple sampling neurons (connected via distinct synapses).
Similarly, in order to allow sign switches during training, connections between sampling neurons were implemented by pairs of synapses (one excitatory and one inhibitory) as well.

The dynamics of the sampling neurons were rendered stochastic in two different ways.
The first setup served as a benchmark and represented a straightforward implementation of the theoretical model from \citep{petrovici2016stochastic}, with Poisson noise generated on the host computer and fed in during the experiment (\cref{fig:networkSetup} A).
In the second setup, we used the spiking activity of a sparse recurrent random network (RN) of inhibitory neurons, instantiated on the same wafer, as a source of noise (\cref{fig:networkSetup} B).
For a more detailed study of sampling-based Bayesian inference with noise generated by deterministic networks, we refer to \citep{jordan2017stochastic}.
The mutual inhibition ensured a relatively constant (sub)population firing rate with suitable random statistics that can replace the ideal Poisson noise in our application.
Projections from the RN to the SSN were chosen as random and sparse; this resulted in weak, but non-zero shared-input correlations.
The remaining correlations are compensated by appropriate training; the Hebbian learning rule (\cref{eq:wake_sleep}) changes the weights and biases in the network such that they cancel the input correlations induced by the RN activity \citep{bytschok2017spike,dold2018stochasticity}.
Hence, the same plasticity rule simultaneously addresses three issues: the learning procedure itself, the compensation of analog variability in neuronal excitability, and the compensation of cross-correlations in the input coming from the background network.
This allowed the hardware-emulated RN to replace the Poisson noise required by the theoretical model.

With these noise-generating mechanisms, the activation function of the neurons, defined by the firing rate as a function of the bias weight $\wb$, took on an approximately logistic shape, as required by the sampling model (\cref{fig:networkSetup} C).
Due mainly to the variability of the hardware circuits {and the precision of the analog parameters}, the exact shape of this activation function varied significantly between neurons (\cref{fig:networkSetup} D-E).
Effectively, this means that initial weights and biases were set randomly, but also that the effective learning rates were different for each neuron.
However, as we show below, this did not prevent the training procedure from converging to a good solution.
This robustness with respect to substrate variability represents an important result of this work.
The used neuron parameters are shown in \cref{table:neuron} and a summary of the used networks is given in \cref{table:netParam}. Our largest experiment a network of 609 neurons with 208 sampling neurons, 1 bias neuron and 400 neurons in the RN (\cref{table:netParam} C) used hardware resources on 28 HICANN chips distributed over 7 reticles. Each of these functional neurons was realized by combining four of the 512 neuronal compartments ("denmems") available on each HICANN, in order to reduce variability in their leak potentials and membrane time constants; for details see \citep{schemmel2010wafer}.

To train the networks on a neuromorphic substrate without embedded plasticity, we used a training concept often referred to as in-the-loop training \citep{schmuker2014neuromorphic,esser2016cover,schmitt2017neuromorphic}.
With the setup discussed above, the only parameters changed during training were digital, namely the synaptic weights between sampling neurons and the weights between bias and sampling neurons.
This allowed us to work with a fixed set of analog parameters, which significantly amplified the precision and speed of reconfiguration during learning, as compared to having used the analog storage instead.
The updates of the digital parameters (synaptic weights) were calculated on the host computer based on the wake-sleep algorithm (\cref{eq:wake_sleep}) but using the spiking activity measured on the hardware.
During the iterative procedure, the values of the weights were saved and updated as a double precision floating point variable, followed by (deterministic) discretization in order to comply with the single-synapse weight resolution of \SI{4}{bits}.
The learning parameters are given in \cref{table:learnParam}.
Clamping (i.e.\ forcing neurons into state 1 or 0 with strong excitatory or inhibitory input) was done by injecting regular spike trains with \SI{100}{Hz} frequency from the host through 5 synapses simultaneously, excitatory for $z_k=1$ and inhibitory for $z_k=0$.
These multapses (multiple synapses connecting two neurons) were needed to exceed the upper limit of single synaptic weights and thus ensure proper clamping.

\section{Results}

\subsection{Learning to approximate a target distribution}
\label{sec:distr}

The experiments described in this section serve as a general benchmark for the ability of our hardware-emulated SSNs and the associated training algorithm to approximate fully specified target Boltzmann distributions.
The viability of our proposal to simultaneously embed deterministic RNs as sources of pseudo-stochasticity is tested by comparing the sampling accuracy of RN-driven SSNs to the case where noise is injected from the host as perfectly uncorrelated Poisson spike trains.

Target distributions $\ptarget$ over 5 RVs were chosen by sampling weights and biases from a Beta distribution centered around zero: $b_i, w_{ji} \sim 2[\mathrm{Beta}(0.5,0.5) - 0.5]$.
Similarly to previous studies \citep{petrovici2016stochastic,jordan2017stochastic}, by giving preference to larger absolute values of the target distribution's parameters, we thereby increased the probability of instantiating rougher, more interesting energy landscapes.
The initial weights and biases of the network were sampled from a uniform distribution over the possible hardware weights.
Due to the small size of the state space, the ``wake'' component of the wake-sleep updates could be calculated analytically as $\langle z_i z_j \rangle = \ptarget(z_i = 1, z_j = 1)$ and $\langle z_i \rangle = \ptarget(z_i=1)$ by explicit marginalization of the target distribution over non-relevant RVs.

\begin{figure}
    \centering
    \includegraphics[width=\textwidth]{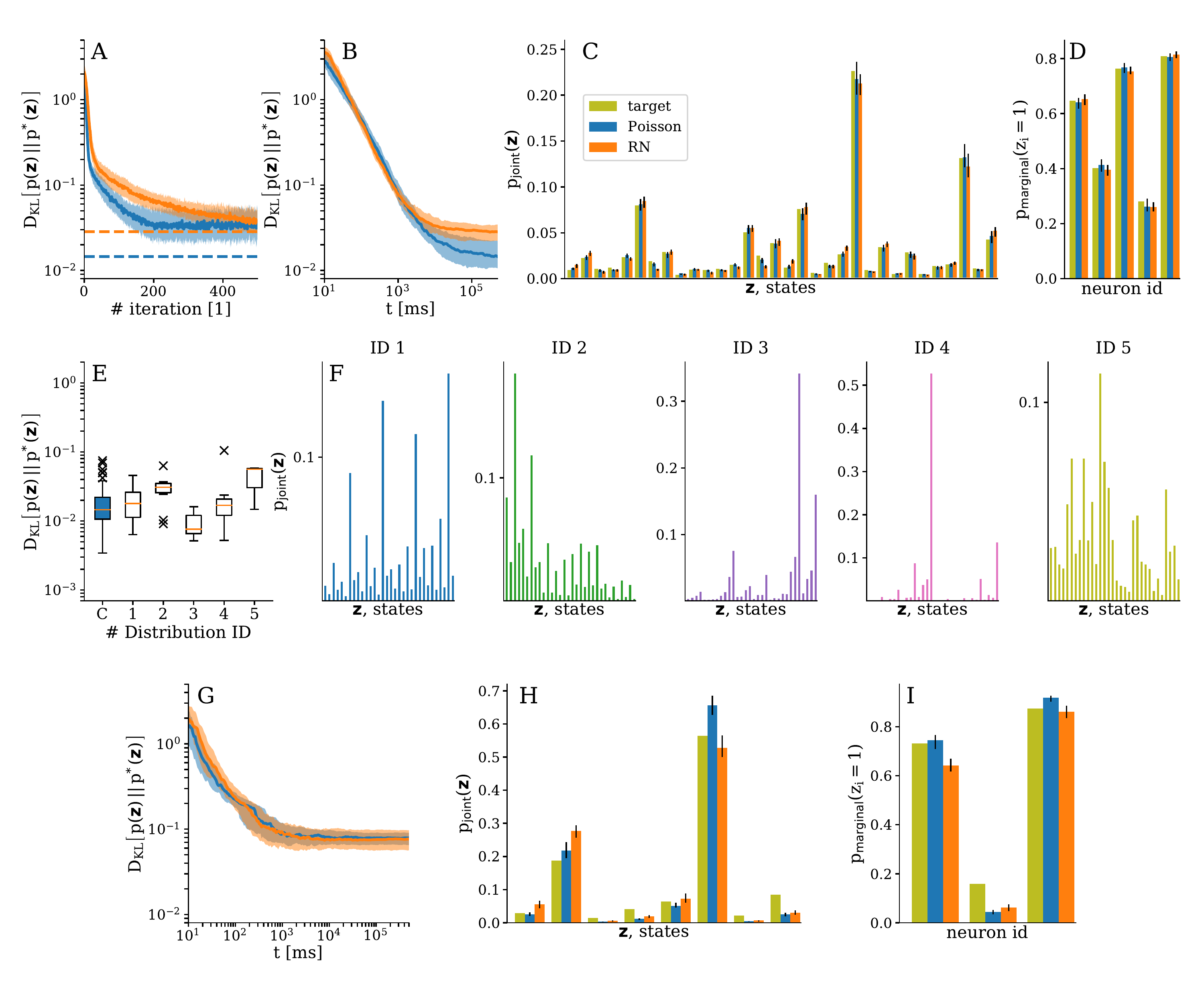}
    \caption{\textbf{Emulated SSNs sampling from target Boltzmann distributions.}
        Sampled distributions are depicted in blue for setups with Poisson noise and in orange for setups using RNs.
        Target distributions shown in dark yellow.
        Data was gathered from 150 runs with random initializations.
        Median values are shown as dark colors and interquartile ranges as either light colors or error bars.
        \textbf{(A)} Improvement of sampled distributions during training.
        The observed variability after convergence (during the plateau) is not due to noise in the system, but rather a consequence of the weight discretization: when the ideal (target) weights lie approximately mid-way between two consecutive integer values on the hardware, training leads to oscillations between these values.
        The parameter configuration showing the best performance during a training run -- which, due to the abovementioned oscillations, was not necessarily the one in the final iteration -- was chosen as the end result of the training phase.
        Averages of these results are shown as dashed lines.
        \textbf{(B)} Convergence of sampled distributions for the trained SSNs.
        \textbf{(C)} and \textbf{(D)} Sampled joint and marginal distributions of the trained SSNs after \SI{5e5}{ms}, respectively.
        \textbf{(E)} Consistency of training results for different target distributions using Poisson noise.
        Here, we show a representative selection of 6 distributions with 10 independent runs per distribution.
        The box highlighted in blue corresponds to the target distribution used in the other panels of \cref{fig:proofOfConcept}.
        {
        The data is plotted following the traditional box-and-whiskers scheme: the orange line represents the median, the box represents the interquartile range, the whiskers represent the full data range and the $\times$ represent the far outliers.
        \textbf{(F)} Target distributions corresponding to the last five box-and-whiskers plots in \textbf{(E)}.}
        \textbf{(G)} Convergence of conditional distributions for the trained SSNs.
        \textbf{(H)} and \textbf{(I)} Sampled conditional joint and marginal distributions of the trained SSNs after \SI{5e5}{ms}, respectively.
        }
    \label{fig:proofOfConcept}
\end{figure}

For training, we used 500 iterations with \SI{1e5}{ms} sampling time per iteration.
Afterwards, the parameter configuration that produced the lowest $\DKL(\psampled \parallel \ptarget)$ was tested in a longer (\SI{5e5}{ms}) experiment.
To study the ability of the trained networks to perform Bayesian inference, we clamped two of the five neurons to fixed values $(z_1, z_2) = (0,1)$ and compared the sampled conditional distribution to the target conditional distribution.
Results for one of these target distributions are shown in \cref{fig:proofOfConcept}.

On average, with Poisson noise, the training showed fast convergence during the first 20 iterations, followed by fine-tuning and full convergence within 200 iterations.
As expected, the convergence of the setups using RNs was significantly slower due to the need to overcome the additional background correlations, but they were still able to achieve similar performance (\cref{fig:proofOfConcept} A).

In both setups, during the test run, the trained SSNs converged to the target distribution following an almost identical power law, which indicates similar mixing properties (\cref{fig:proofOfConcept} B).
For longer sampling durations ($\gg\SI{10e3}{ms}$), the systematic deviations from the target distributions become visible and the $\DKL(\psampled \parallel \ptarget)$ reaches the same plateau at approximately $\DKL(\psampled \parallel \ptarget) \approx \num{2e-2}$ as observed during training.
\Cref{fig:proofOfConcept} C and D respectively show the sampled joint and marginal distributions after convergence.
These observations remained consistent across a set of 20 different target distributions (see \cref{fig:proofOfConcept} E for a representative selection).

Similar observations hold for the inference experiments.
Due to the smaller state space, convergence happened faster (\cref{fig:proofOfConcept} E).
The corresponding joint and marginal distributions are shown in \cref{fig:proofOfConcept} F and G, respectively.
The lower accuracy of these distributions is mainly because of the asymmetry of the effective synaptic weights caused by the variability of the substrate, towards which the learning algorithm is agnostic.
The training took \SI{5e2}{s} wall-clock time, including the pure experiment runtime, the initialization of the hardware and the calculation of the updates on the host computer (total turn-over time of the training).
This corresponds to a speed-up factor of 100 compared to the equivalent \SI{5e4}{s} of biological real time.
While the nominal $10^4$ speed-up remained intact for the emulation of network dynamics, the total speed-up factor was reduced due to the overhead imposed by network (re)configuration and I/O between the host and the neuromorphic substrate.

\begin{figure}
    \centering
    \includegraphics[width=\textwidth]{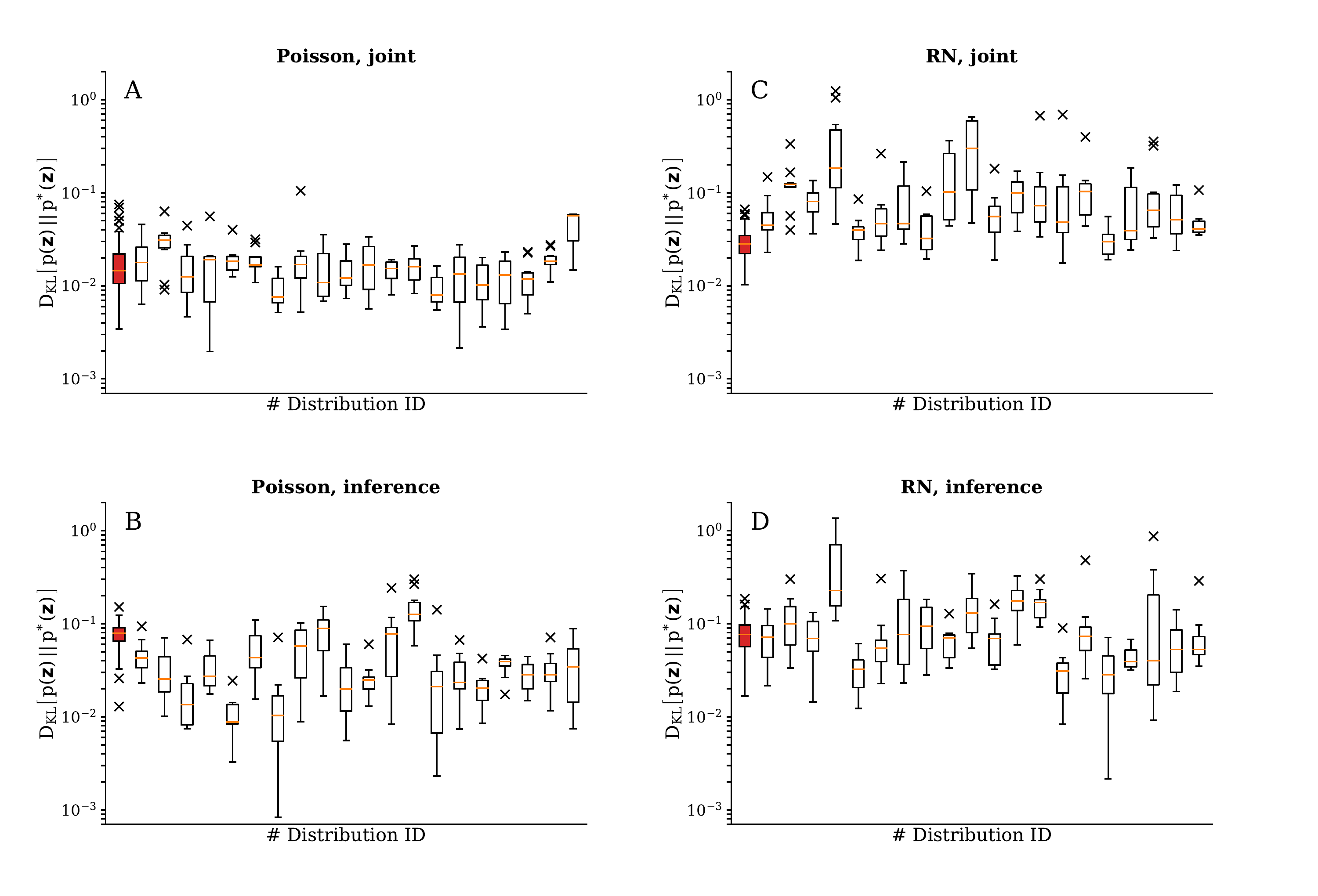}
    \caption{
        \textbf{Emulated SSNs sampling from different target Boltzmann distributions.}
        The figure shows the results of experiments identical to the ones in \cref{sec:distr} for 20 different target distributions with 10 repetitions for each sample.
        We show the $\DKL(\psampled \parallel \ptarget)$ of the test-run after training for \textbf{(A)} the joint distributions with Poisson noise, \textbf{(B)} the inference experiment with Poisson noise, \textbf{(C)} the joint distributions with a random background network and \textbf{(C)} the inference experiment with a random background network.
        The data is plotted following the traditional box-and-whiskers scheme: the orange line represents the median, the box represents the interquartile range, the whiskers represent the full data range and the $\times$ represent the far outliers.
        In each subplot the leftmost data (highlighted in red) corresponds to the distribution shown in \cref{fig:proofOfConcept}.
        }
    \label{fig:checkTargets}
\end{figure}

We carried out the same experiments as described previously with 20 different samples for the weights and the biases of the target distribution.
In \cref{fig:checkTargets} we show the final DKLs after training to represent a target distribution both with Poisson noise and with the activity of a random network.
The experiments were repeated 10 times for each sample.
Median learning results remained consistent across target distributions, with the variability reflecting the difficulty of the problem (discrepancies between LIF and Glauber dynamics become more pronounced for larger weights and biases).
Variability across trials for the same target distribution is due to the trial-to-trial variability of the analog parameter storage (floating gates), due to the inherent stochasticity in the learning procedure (sampling accuracy in an update step), as well as due to systematic discrepancies between the effective pre-post and post-pre interaction strengths between sampling units, which are themselves a consequence of the aforementioned floating gate variability.

\subsection{Learning from data}
\label{sec:datasets}

\begin{figure}
    \includegraphics[width=\textwidth]{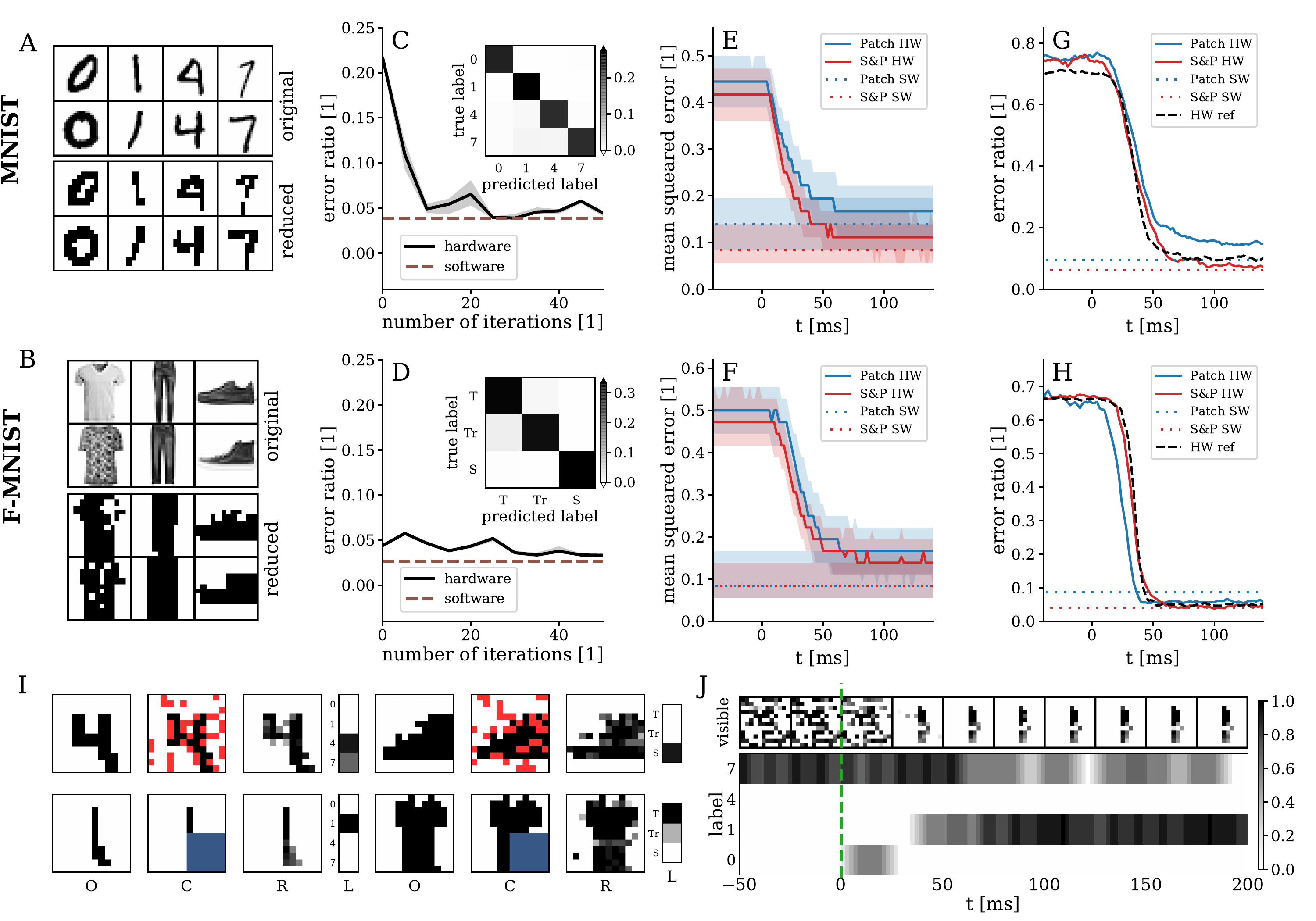}
    \caption{
        \textbf{Behavior of hierarchical SSNs trained on data.}
        Top row: rMNIST; middle row: rFMNIST; bottom row: exemplary setups for the partial occlusion scenarios.
        \textbf{(A-B)} Exemplary images from the rMNIST (A) and rFMNIST (B) datasets used for training and comparison to their MNIST and FMNIST originals.
        \textbf{(C-D)} Training with the hardware in the loop after translation of pre-trained parameters.
        Confusion matrices after training shown as insets.
        Performance of the reference RBMs shown as dashed brown lines.
        Results are given as median and interquartile values over 10 test runs.
        \textbf{(E-F)} Pattern completion and \textbf{(G-H)} error ratio of the inferred label for partially occluded images (blue: patch; red: salt\&pepper).
        Solid lines represent median values and shaded areas show interquartile ranges over 250 test images per class.
        Performance of the reference RBMs shown as dashed lines.
        As a reference, we also show the error ratio of the SNNs on unoccluded images in (G) and (H).
        \textbf{(I)} Snapshots of the pattern completion experiments: O - original image, C - clamped image (red and blue pixels are occluded), R - response of the visible layer, L - response of the label layer.
        \textbf{(J)} Exemplary temporal evolution of a pattern completion experiment with patch occlusion.
        For better visualization of the activity in the visible layer in (J) and (I), we smoothed out its discretized response to obtain grayscale pixel values, by convolving its state vector with a box filter of \SI{10}{\milli\second} width.
        }
    \label{fig:datasets}
\end{figure}

In order to obtain models of labeled data, we trained hierarchical SSNs analogously to restricted Boltzmann machines (RBMs).
Here, we used two different datasets: a reduced version of the MNIST~\citep{lecun1998gradient} and the fashion MNIST~\citep{xiao2017online} datasets, which we abbreviate as rMNIST and rFMNIST in the following.
The images were first reduced with nearest-neighbor resampling (\texttt{misc.imresize} function in the SciPy library~\citep{jones2014scipy}) and then binarized around the median gray value over each image.
We used all images from the original datasets (approx. 6000 per class) from 4 classes (0, 1, 4, 7) for rMNIST and 3 classes (\textbf{T}-shirts, \textbf{Tr}ousers, \textbf{S}neakers) for rFMNIST (\cref{fig:datasets} A-B).
The emulated SSNs consisted of 3 layers, with 144 visible, 60 hidden and either 4 label units for rMNIST or 3 for rFMNIST.

Pre-training was done on simulated classical RBMs using the CAST algorithm~\citep{salakhutdinov2010learning}.
The pre-training provided a starting point for training on the hardware in order to accelerate the convergence of the in-the-loop training procedure.
We use the performance of these RBMs in software simulations using Gibbs sampling as a reference for the results obtained with the hardware-emulated SSNs.
After pre-training, we mapped these RBMs to approximately equivalent SSNs on the hardware, using an empirical translation factor based on an average activation function (\cref{fig:networkSetup} C) to calculate the initial hardware synaptic weights from weights and biases of the RBMs.
Especially for rMNIST, this resulted in a significant deterioration of the classification performance (\cref{fig:datasets} C).
After mapping, we continued training using the wake-sleep algorithm, with the hardware in the loop.
While in the previous task it was possible to calculate the data term explicitly, it now had to be sampled as well.
In order to ensure proper clamping, the synapses from the hidden to the label layer and from the hidden layer to the visible layer were turned off during the wake phase.

The SSNs were tested for both their discriminative and their generative properties.
For classification, the visible layer was clamped to images from the test set (black pixels correspond to $z_k=1$ and white pixels to $z_k=0$).
Each image was presented for 500 biological milliseconds, which corresponds to \SI{50}{\mu s} wall-clock time.
The neuron in the label layer with the highest firing rate was interpreted as the label predicted by the model. The spiking activity of the neurons was read out directly from the hardware, without additional off-chip post-processing.
For both datasets, training was able to restore the performance lost in the translation of the abstract RBM to the hardware-emulated SSN.
The emulated SSNs achieved error rates of $4.45^{+0.12}_{-0.36}\%$ on rMNIST and $3.32^{+0.27}_{-0.04}\%$ on rFMNIST.
These values are close to the ones obtained by the reference RBMs: $3.89^{+0.10}_{-0.02}\%$ on rMNIST and $2.645^{+0.002}_{-0.010}\%$ on rFMNIST (\cref{fig:datasets} C-D, confusion matrices shown as insets).

The gross wall-clock time needed to classify the 4125 images in the rMNIST test set was \SI{10}{s} (\SI{2.4}{ms} per image, $210\times$ speed-up).
For the 3000 images in the rFMNIST test set, the emulation ran for \SI{9.4}{s} (\SI{3.1}{ms} per image; $160\times$ speed-up).
This subsumes the runtime of the BrainScaleS software stack, hardware configuration and the network emulation.
The runtime of the software-stack includes the translation from a PyNN-based network description to a corresponding hardware configuration.
As before, the difference between the nominal acceleration factor and the effective speed-up stems from the I/O and initialization overhead of the hardware system.

To test the generative properties of our emulated SSNs, we set up two scenarios requiring them to perform pattern completion.
For each class, 250 incomplete images were presented as inputs to the visible layer.
For each image, \SI{25}{\percent} of visible neurons received no input, with the occlusion following two different schemes: salt\&pepper (upper row in \cref{fig:datasets} I) and patch (lower row in \cref{fig:datasets} I).
Each image was presented for \SI{500}{ms}.
In order to remove any initialization bias resulting from preceding images, random input was applied to the visible layer between consecutive images.

Reconstruction accuracy was measured using the mean squared error (MSE) between the reconstructed and original occluded pixels.
For binary images, as in our case, the MSE reflects the average ratio of mis-reconstructed to total reconstructed pixels.
Simultaneously, we also recorded the classification accuracy on the partially occluded images.
After stimulus onset, the MSE converged from chance level ($\approx \SI{50}{\percent}$) to its minimum ($\approx \SI{10}{\percent}$) within \SI{50}{ms} (\cref{fig:datasets} E-F).
Given an average refractory period of $\approx\SI{10}{ms}$ (\cref{fig:networkSetup} C), this suggests that the network was able to react to the input with no more than 5 spikes per neuron.
For all studied scenarios, the reconstruction performance of the emulated SSNs closely matched the one achieved by the reference RBMs.
Examples of image reconstruction are shown in \cref{fig:datasets} I-J for both datasets and occlusion scenarios.
The classification performance deteriorated only slightly compared to non-occluded images and also remained close to the performance of the reference RBMs (\cref{fig:datasets} G-H).
The temporal evolution of the classification error closely followed that of the MSE.

\begin{figure}
    \includegraphics[width=\textwidth]{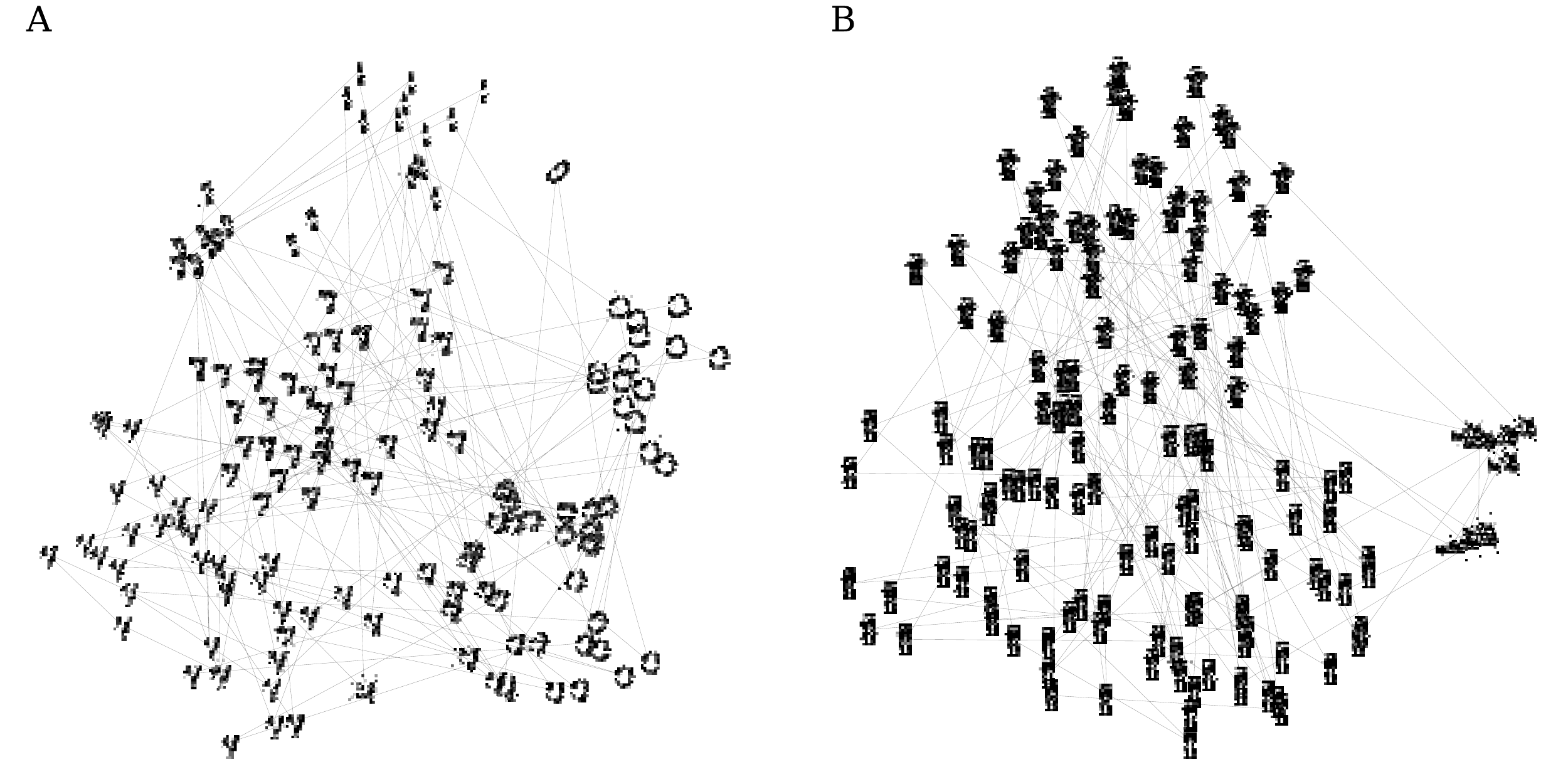}
    \caption{
        \textbf{Generated images during guided dreaming.}
        The visible state space, along with the position of the generated images within it, was projected to two dimensions using t-SNE \citep{maaten2008visualizing}.
        The thin lines connect consecutive samples.
        \textbf{(A)} rMNIST; \textbf{(B)} rFMNIST.
        }
    \label{fig:tsne}
\end{figure}

As a further test of the generative abilities of our hardware-emulated SSNs, we recorded the images produced by the visible layer during guided dreaming.
In this task, the visible and hidden layers of the SSN evolved freely without external input, while the label layer was periodically clamped with external input such that exactly one of the label neurons was active at any time (enforced one-hot coding).
In a perfect model, this would cause the visible layer to sample only from configurations compatible with the hidden layer, i.e., from images corresponding to that particular class.
Between the clamping of consecutive labels, we injected \si{100}{ms} random input to visible layer to facilitate the changing of the image.
The SSNs were able to generate varied and recognizable pictures, within the limits imposed by the low resolution of the visible layer (\cref{fig:tsne}).
For rMNIST, all used classes appeared in correct correspondence to the clamped label.
For rFMNIST, images from the class ``Sneakers'' were not always triggered by the corresponding guidance from the label layer, suggesting that the learned modes in the energy landscape are too deep, and sneakers too dissimilar to T-shirts and Trousers, to allow good mixing during guided dreaming.

\section{Discussion}

This manuscript presents the first scalable demonstration of sampling-based probabilistic inference with spiking networks on a highly accelerated analog neuromorphic substrate.
We trained fully connected spiking networks to sample from target distributions and hierarchical spiking networks as discriminative and generative models of higher-dimensional input data.
Despite the inherent variability of the analog substrate, we were able to achieve performance levels comparable to those of software simulations in several benchmark tasks, while maintaining a significant overall acceleration factor compared to systems that operate in biological real time.
Importantly, by co-embedding the generation of stochasticity within the same substrate, we demonstrated the viability of a fully embedded neural sampling model with significantly reduced demands on off-substrate I/O bandwidth.
Having a fully embedded implementation allows the runtime of the experiments to scale as $\mathcal{O}(1)$ with the size of the emulated network; this is inherent to the nature of physical emulation, for which wall-clock runtime only depends on the emulated time in the biological reference frame.
In the following, we address the limitations of our study, point out links to related work and discuss its implications within the greater context of computational neuroscience and bio-inspired AI.

\subsection{Limitations and constraints}

The most notable limitation imposed by the current commissioning state of the BrainScaleS system was on the size of the emulated SSNs.
At the time of writing, {due to limited software flexibility, system assembly and substrate yield, the usable hardware real-estate was reduced to a patchy and non-contiguous area}, thereby strongly limiting the maximum connectivity between different locations within this area.
In order to limit synapse loss to small values (below \SI{2}{\percent}), we restricted ourselves to using a small but contiguous functioning area of the wafer, which in turn limited the maximum size of our SSNs and noise-generating RNs.
Ongoing improvements in post-production and assembly, as well as in the mapping and routing software, are expected to enhance on-wafer connectivity and thereby automatically increase the size of emulable networks, as the architecture of our SSNs scales naturally to such an increase in hardware resources.

To a lesser extent, the sampling accuracy was also affected by the limited precision of hardware parameter control.
The writing of analog parameters exhibits significant trial-to-trial variability; in any given trial, this leads to a heterogeneous substrate, which is known to reduce the sampling accuracy \citep{probst2015probabilistic}.
Most of this variability is compensated during learning, but the \SI{4}{bit} resolution of the synaptic weights and the imperfect symmetry in the effective weight matrix due to analog variability of the synaptic circuits ultimately limit the ability of the SSN to approximate target distributions.
This leads to the ``jumping'' behavior of the $\DKL(\psampled \parallel \ptarget)$ in the final stages of learning (\cref{fig:proofOfConcept} A).
In smaller networks, synaptic weight resolution is a critical performance modifier \citep{petrovici2017robustness}.
However, the penalty imposed by a limited synaptic weight resolution is known to decrease for larger deep networks with more and larger hidden layers, both spiking and non-spiking \citep{courbariaux2015binaryconnect,petrovici2017pattern}.
Furthermore, the successor system \citep[BrainScaleS-2, ][]{aamir2016highly} is designed with a 6-bit weight resolution.

{
In the setup we used shared bias neurons for several neurons in the sampling network.
This helped us save hardware resources, thus allowing the emulation of larger functional networks.
Such bias neuron sharing is expected to introduce some small amount of temporal correlations between the sampling neurons.
However, this effect was too small to observe in our experiments for several reasons.
First, the high firing rate of the bias neurons helped smooth out the bias voltage induced into the sampling neurons.
Second, the different delays and spike timing jitter on the hardware reduces such cross-correlations.
Third, other dominant limitations overshadow the effect of shared bias neurons.
In any case, the used training procedure inherently compensates for excess cross-correlations, thus effectively removing any distortions to the target distribution that this effect might introduce \citep{bytschok2017spike,dold2018stochasticity}.
}

In the current setup, our SSNs displayed limited mixing abilities.
During guided dreaming, images from one of the learned classes were more difficult to generate (\cref{fig:tsne}).
Restricted mixing due to deep modes in the energy landscape carved out by contrastive learning is a well-known problem for classical Boltzmann machines, which is usually alleviated by computationally costly annealing techniques \citep{salakhutdinov2010learning,desjardins2010parallel,bengio2013better}.
However, the fully-commissioned BrainScaleS system will feature embedded short-term synaptic plasticity \citep{schemmel2010wafer}, which has been shown to promote mixing in spiking networks \citep{leng2018spiking} while operating purely locally, at the level of individual synapses.

Currently, the execution speed of emulation runs is dominated by the I/O overhead, which in turn is mostly spent on setting up the experiment. This leads to the classification (\cref{sec:datasets}) of one image taking \SIrange{2.4}{3.9}{ms}, whereas the pure network runtime is merely $\SI{50}{\mu s}$. A streamlining of the software layer that performs this setup is expected to significantly reduce this discrepancy.

The synaptic learning rule was local and Hebbian, but updates were calculated on a host computer using an iterative in-the-loop training procedure, which required repeated stopping, evaluation and restart of the emulation, thereby reducing the nominal acceleration factor of $10^4$ by two orders of magnitude.
By utilizing on-chip plasticity, as available, for example, on the BrainScaleS-2 successor system \citep{friedmann2017demonstrating,wunderlich2018demonstrating}, this laborious procedure becomes obsolete and the accelerated nature of the substrate can be exploited to its fullest extent.

\subsection{Relation to other work}

This study builds upon a series of theoretical and experimental studies of sampling-based probabilistic inference using the dynamics of biological neurons.
The inclusion of refractory times was first considered in \citep{buesing2011neural}.
An extension to networks of leaky integrate-and-fire neurons and a theoretical framework for their dynamics and statistics followed in \citep{petrovici2013stochastic} and \citep{petrovici2016stochastic}.
The compensation of shared-input correlations through inhibitory feedback and learning was discussed in \citep{jordan2017stochastic}, \citep{bytschok2017spike} and \citep{dold2018stochasticity}, inspired by the early study of asynchronous irregular firing in \citep{brunel2000dynamics} and by preceding correlation studies in theoretical \citep{tetzlaff2012decorrelation} and experimental \citep{pfeil2016effect} work.

Previous small-scale studies of sampling on accelerated mixed-signal neuromorphic hardware include \citep{petrovici2015sampling,petrovici2017robustness,petrovici2017pattern}.
An implementation of sampling with spiking neurons and its application to the MNIST dataset was shown in \citep{pedroni2016mapping} using the fully digital, real-time TrueNorth neuromorphic chip \citep{merolla2014million}.

We stress two important differences between \citep{pedroni2016mapping} and this work.
First, the nature of the neuromorphic substrate: the TrueNorth system is fully digital and calculates neuronal state updates numerically, in contrast to the physical-model paradigm instantiated by BrainScaleS.
In this sense, TrueNorth emulations are significantly closer to classical computer simulations on parallel machines: updates of dynamical variables are precise and robustness to variability is not an issue; { however TrueNorth typically runs in biological real time \citep{merolla2014million,akopyan2015truenorth},} which is 10.000 times slower than BrainScaleS.
Second, the nature of neuron dynamics: the neuron model used in \citep{pedroni2016mapping} is an intrinsically stochastic unit that sums its weighted inputs, thus remaining very close to classical Gibbs sampling and Boltzmann machines, while our approach considers multiple additional aspects of its biological archetype (exponential synaptic kernels, leaky membranes, deterministic firing, stochasticity through synaptic background, shared-input correlations etc.).
Moreover, our approach uses fewer hardware neuron units to represent a sampling unit, enabling a more parsimonious utilization of the neuromorphic substrate.

\subsection{Conclusion}

In this work we showed how sampling-based Bayesian inference using hierarchical spiking networks can be robustly implemented on a physical model system despite inherent variability and imperfections.
Underlying neuron and synapse dynamics are deterministic and close to their biological archetypes, but with much shorter time constants, hence the intrinsic acceleration factor of $10^4$ with respect to biology.
The entire architecture -- sampling network plus background random network -- was fully deterministic and entirely contained on the neuromorphic substrate, with external communication used only to represent input patterns and labels.
Considering the deterministic nature of neurons in vitro \citep{mainen1995reliability, reinagel2002precise, toups2012multiple}, such an architecture also represents a plausible model for neural sampling in cortex { \citep{jordan2017stochastic,dold2018stochasticity}.}

We demonstrated sampling from arbitrary Boltzmann distributions over binary random variables, as well as generative and discriminative properties of networks trained with visual data.
The framework can be extended to sampling from arbitrary probability distributions over binary random variables, as it was shown in software simulations \citep{probst2015probabilistic}.
For such networks, the two abovementioned computational tasks (pattern completion and classification) happen simultaneously, as they both require the calculation of conditional distributions, which is carried out implicitly by the network dynamics.
Both during learning and for the subsequent inference tasks, the setup benefitted significantly from the fast intrinsic dynamics of the substrate, achieving a net speedup of \numrange{100}{210} compared to biology.

We view these results as a contribution to the nascent, but expanding field of applications for biologically inspired physical-model systems.
They demonstrate the feasibility of such devices for solving problems in machine learning, as well as for studying biological phenomena.
Importantly, they explicitly addresses the search for robust computational models that are able to harness the strengths of these systems, most importantly their speed and energy efficiency.
The proposed architecture scales naturally to substrates with more neuronal real-estate and can be used for a wide array of tasks that can be mapped to a Bayesian formulation, such as constraint satisfaction problems \citep{jonke2016solving,fonseca2017using}, prediction of temporal sequences \citep{sutskever2007learning}, movement planning \citep{taylor2009factored,alemi2015affect}, simulation of solid-state systems \citep{edwards1975theory} and quantum many-body problems \citep{carleo2017solving,czischek2018quenches}.

\section*{Conflict of Interest Statement}
The authors declare that the research was conducted in the absence of any commercial or financial relationships that could be construed as a potential conflict of interest.

\section*{Author Contributions}
Akos F. Kungl (AFK), Andreas Baumbach (AB), Dominik Dold (DD), Luziwei Leng (LL), Sebastian Schmitt (SS), Paul M\"uller (PM) and Mihai A. Petrovici (MAP) designed the study.
AFK conducted the experiments and the evaluations.
Nico G\"urtler (NG) contributed to the evaluations and provided software support for the evaluation.
AFK wrote the initial manuscript.
Eric M\"uller (ECM), Christian Mauch (CM), Johann Kl\"ahn (JK), Sebastian Schmitt (SS), Kai Husmann (KH) and Oliver Breitwieser (OB) supported experiment realization; ECM coordinated the software development for the neuromorphic systems.
Alexander Kugele (AK), Christoph Koke (CK) and Mitja Kleider (MK) contributed with characterization, calibration testing and debugging of the system.
Andreas Gr\"ubl (AG), Dan Husmann (DH), Maurice G\"uttler (MG) were responsible for system assembly.
AG did the digital front- and back-end implementation.
Vitali Karasenko (VK) provided FPGA firmware and supported system commissioning.
Johannes Schemmel (JS) is the architect and lead designer of the neuromorphic platform.
MAP, Karlheinz Meier (KM), JS, SS and ECM provided conceptual and scientific advice.
All authors contributed to the final manuscript.

\section*{Funding}
The work leading to these results has received funding from the European Union Seventh Framework Programme (FP7) under grant agreement No \#604102, the EU's Horizon 2020 research and innovation programme under grant agreements No \#720270 and \#785907 (Human Brain Project, HBP), the EU's research project BrainScaleS \#269921 and the Heidelberg Graduate School of Fundamental Physics.
We acknowledge financial support by Deutsche Forschungsgemeinschaft within the funding programme Open Access Publishing, by the Baden-Württemberg Ministry of Science, Research and the Arts and by Ruprecht-Karls-Universität Heidelberg.
We owe particular gratitude to the sustained support of our research by the Manfred St\"ark Foundation.

\section*{Acknowledgments}
We thank Johannes Bill for many fruitful discussions.

\section*{Supplemental Data}
Two videos can be found in the online version of the article that exemplify fully-embedded sampling from a target distribution and fully-embedded pattern completion on the BrainScaleS-1.

\bibliographystyle{frontiersinSCNS_ENG_HUMS} 
\bibliography{bib}

\begin{thebibliography}{74}
\providecommand{\natexlab}[1]{#1}
\expandafter\ifx\csname urlstyle\endcsname\relax
  \providecommand{\doi}[1]{doi:\discretionary{}{}{}#1}\else
  \providecommand{\doi}{doi:\discretionary{}{}{}\begingroup
  \urlstyle{rm}\Url}\fi
\providecommand{\selectlanguage}[1]{\relax}
\providecommand{\bibAnnoteFile}[1]{%
  \IfFileExists{#1}{\begin{quotation}\noindent\textsc{Key:} #1\\
  \textsc{Annotation:}\ \input{#1}\end{quotation}}{}}
\providecommand{\bibAnnote}[2]{%
  \begin{quotation}\noindent\textsc{Key:} #1\\
  \textsc{Annotation:}\ #2\end{quotation}}

\bibitem[{Aamir et~al.(2016)Aamir, M{\"u}ller, Hartel, Schemmel, and
  Meier}]{aamir2016highly}
Aamir, S.~A., M{\"u}ller, P., Hartel, A., Schemmel, J., and Meier, K. (2016).
\newblock A highly tunable 65-nm cmos lif neuron for a large scale neuromorphic
  system.
\newblock In \emph{European Solid-State Circuits Conference, ESSCIRC Conference
  2016: 42nd} (IEEE), 71--74
\bibAnnoteFile{aamir2016highly}

\bibitem[{Aitchison and Lengyel(2016)}]{aitchison2016hamiltonian}
Aitchison, L. and Lengyel, M. (2016).
\newblock The hamiltonian brain: efficient probabilistic inference with
  excitatory-inhibitory neural circuit dynamics.
\newblock \emph{PLoS computational biology} 12, e1005186
\bibAnnoteFile{aitchison2016hamiltonian}

\bibitem[{Akopyan et~al.(2015)Akopyan, Sawada, Cassidy, Alvarez-Icaza, Arthur,
  Merolla et~al.}]{akopyan2015truenorth}
Akopyan, F., Sawada, J., Cassidy, A., Alvarez-Icaza, R., Arthur, J., Merolla,
  P., et~al. (2015).
\newblock Truenorth: Design and tool flow of a 65 mw 1 million neuron
  programmable neurosynaptic chip.
\newblock \emph{IEEE Transactions on Computer-Aided Design of Integrated
  Circuits and Systems} 34, 1537--1557
\bibAnnoteFile{akopyan2015truenorth}

\bibitem[{Alemi et~al.(2015)Alemi, Li, and Pasquier}]{alemi2015affect}
Alemi, O., Li, W., and Pasquier, P. (2015).
\newblock Affect-expressive movement generation with factored conditional
  restricted boltzmann machines.
\newblock In \emph{2015 International Conference on Affective Computing and
  Intelligent Interaction (ACII)} (IEEE), 442--448
\bibAnnoteFile{alemi2015affect}

\bibitem[{Bengio et~al.(2013)Bengio, Mesnil, Dauphin, and
  Rifai}]{bengio2013better}
Bengio, Y., Mesnil, G., Dauphin, Y., and Rifai, S. (2013).
\newblock Better mixing via deep representations.
\newblock In \emph{International conference on machine learning}. 552--560
\bibAnnoteFile{bengio2013better}

\bibitem[{Berkes et~al.(2011)Berkes, Orb{\'a}n, Lengyel, and
  Fiser}]{berkes2011spontaneous}
Berkes, P., Orb{\'a}n, G., Lengyel, M., and Fiser, J. (2011).
\newblock Spontaneous cortical activity reveals hallmarks of an optimal
  internal model of the environment.
\newblock \emph{Science} 331, 83--87
\bibAnnoteFile{berkes2011spontaneous}

\bibitem[{Brette and Gerstner(2005)}]{brette2005adaptive}
Brette, R. and Gerstner, W. (2005).
\newblock Adaptive exponential integrate-and-fire model as an effective
  description of neuronal activity.
\newblock \emph{Journal of neurophysiology} 94, 3637--3642
\bibAnnoteFile{brette2005adaptive}

\bibitem[{Brunel(2000)}]{brunel2000dynamics}
Brunel, N. (2000).
\newblock Dynamics of sparsely connected networks of excitatory and inhibitory
  spiking neurons.
\newblock \emph{Journal of computational neuroscience} 8, 183--208
\bibAnnoteFile{brunel2000dynamics}

\bibitem[{Buesing et~al.(2011)Buesing, Bill, Nessler, and
  Maass}]{buesing2011neural}
Buesing, L., Bill, J., Nessler, B., and Maass, W. (2011).
\newblock Neural dynamics as sampling: a model for stochastic computation in
  recurrent networks of spiking neurons.
\newblock \emph{PLoS computational biology} 7, e1002211
\bibAnnoteFile{buesing2011neural}

\bibitem[{Bytschok et~al.(2017)Bytschok, Dold, Schemmel, Meier, and
  Petrovici}]{bytschok2017spike}
Bytschok, I., Dold, D., Schemmel, J., Meier, K., and Petrovici, M.~A. (2017).
\newblock Spike-based probabilistic inference with correlated noise.
\newblock In \emph{BMC Neuroscience 2017} (Organization for Computational
  Neurosciences), vol.~18, P200
\bibAnnoteFile{bytschok2017spike}

\bibitem[{Carleo and Troyer(2017)}]{carleo2017solving}
Carleo, G. and Troyer, M. (2017).
\newblock Solving the quantum many-body problem with artificial neural
  networks.
\newblock \emph{Science} 355, 602--606
\bibAnnoteFile{carleo2017solving}

\bibitem[{Chang et~al.(2016)Chang, Fowler, Chen, Zhou, Pan, Chang
  et~al.}]{chang2016demonstration}
Chang, Y.-F., Fowler, B., Chen, Y.-C., Zhou, F., Pan, C.-H., Chang, T.-C.,
  et~al. (2016).
\newblock Demonstration of synaptic behaviors and resistive switching
  characterizations by proton exchange reactions in silicon oxide.
\newblock \emph{Scientific reports} 6, 21268
\bibAnnoteFile{chang2016demonstration}

\bibitem[{Courbariaux et~al.(2015)Courbariaux, Bengio, and
  David}]{courbariaux2015binaryconnect}
Courbariaux, M., Bengio, Y., and David, J.-P. (2015).
\newblock Binaryconnect: Training deep neural networks with binary weights
  during propagations.
\newblock In \emph{Advances in neural information processing systems}.
  3123--3131
\bibAnnoteFile{courbariaux2015binaryconnect}

\bibitem[{Czischek et~al.(2018)Czischek, G{\"a}rttner, and
  Gasenzer}]{czischek2018quenches}
Czischek, S., G{\"a}rttner, M., and Gasenzer, T. (2018).
\newblock Quenches near ising quantum criticality as a challenge for artificial
  neural networks.
\newblock \emph{Physical Review B} 98, 024311
\bibAnnoteFile{czischek2018quenches}

\bibitem[{Davison et~al.(2009)Davison, Br{\"u}derle, Eppler, Kremkow, Muller,
  Pecevski et~al.}]{davison2009pynn}
Davison, A.~P., Br{\"u}derle, D., Eppler, J.~M., Kremkow, J., Muller, E.,
  Pecevski, D., et~al. (2009).
\newblock Pynn: a common interface for neuronal network simulators.
\newblock \emph{Frontiers in neuroinformatics} 2, 11
\bibAnnoteFile{davison2009pynn}

\bibitem[{Desjardins et~al.(2010)Desjardins, Courville, Bengio, Vincent, and
  Delalleau}]{desjardins2010parallel}
Desjardins, G., Courville, A., Bengio, Y., Vincent, P., and Delalleau, O.
  (2010).
\newblock Parallel tempering for training of restricted boltzmann machines.
\newblock In \emph{Proceedings of the thirteenth international conference on
  artificial intelligence and statistics} (MIT Press Cambridge, MA), 145--152
\bibAnnoteFile{desjardins2010parallel}

\bibitem[{Destexhe et~al.(2003)Destexhe, Rudolph, and
  Par{\'e}}]{destexhe2003high}
Destexhe, A., Rudolph, M., and Par{\'e}, D. (2003).
\newblock The high-conductance state of neocortical neurons in vivo.
\newblock \emph{Nature reviews neuroscience} 4, 739
\bibAnnoteFile{destexhe2003high}

\bibitem[{Dold et~al.(2019)Dold, Bytschok, Kungl, Baumbach, Breitwieser, Senn
  et~al.}]{dold2018stochasticity}
Dold, D., Bytschok, I., Kungl, A.~F., Baumbach, A., Breitwieser, O., Senn, W.,
  et~al. (2019).
\newblock Stochasticity from function --- why the bayesian brain may need no
  noise.
\newblock \emph{Neural Networks} 119, 200--213
\bibAnnoteFile{dold2018stochasticity}

\bibitem[{Edwards and Anderson(1975)}]{edwards1975theory}
Edwards, S.~F. and Anderson, P.~W. (1975).
\newblock Theory of spin glasses.
\newblock \emph{Journal of Physics F: Metal Physics} 5, 965
\bibAnnoteFile{edwards1975theory}

\bibitem[{Esser et~al.(2016)Esser, Merolla, Arthur, Cassidy, Appuswamy,
  Andreopoulos et~al.}]{esser2016cover}
Esser, S.~K., Merolla, P.~A., Arthur, J.~V., Cassidy, A.~S., Appuswamy, R.,
  Andreopoulos, A., et~al. (2016).
\newblock From the cover: Convolutional networks for fast, energy-efficient
  neuromorphic computing.
\newblock \emph{Proceedings of the National Academy of Sciences of the United
  States of America} 113, 11441
\bibAnnoteFile{esser2016cover}

\bibitem[{Fonseca~Guerra and Furber(2017)}]{fonseca2017using}
Fonseca~Guerra, G.~A. and Furber, S.~B. (2017).
\newblock Using stochastic spiking neural networks on spinnaker to solve
  constraint satisfaction problems.
\newblock \emph{Frontiers in neuroscience} 11, 714
\bibAnnoteFile{fonseca2017using}

\bibitem[{Friedmann et~al.(2017)Friedmann, Schemmel, Gr{\"u}bl, Hartel, Hock,
  and Meier}]{friedmann2017demonstrating}
Friedmann, S., Schemmel, J., Gr{\"u}bl, A., Hartel, A., Hock, M., and Meier, K.
  (2017).
\newblock Demonstrating hybrid learning in a flexible neuromorphic hardware
  system.
\newblock \emph{IEEE transactions on biomedical circuits and systems} 11,
  128--142
\bibAnnoteFile{friedmann2017demonstrating}

\bibitem[{Furber(2016)}]{furber2016large}
Furber, S. (2016).
\newblock Large-scale neuromorphic computing systems.
\newblock \emph{Journal of neural engineering} 13, 051001
\bibAnnoteFile{furber2016large}

\bibitem[{Haefner et~al.(2016)Haefner, Berkes, and
  Fiser}]{haefner2016perceptual}
Haefner, R.~M., Berkes, P., and Fiser, J. (2016).
\newblock Perceptual decision-making as probabilistic inference by neural
  sampling.
\newblock \emph{Neuron} 90, 649--660
\bibAnnoteFile{haefner2016perceptual}

\bibitem[{Hennequin et~al.(2014)Hennequin, Aitchison, and
  Lengyel}]{hennequin2014fast}
Hennequin, G., Aitchison, L., and Lengyel, M. (2014).
\newblock Fast sampling for bayesian inference in neural circuits.
\newblock \emph{arXiv preprint arXiv:1404.3521}
\bibAnnoteFile{hennequin2014fast}

\bibitem[{Hinton(2012)}]{hinton2012practical}
Hinton, G.~E. (2012).
\newblock A practical guide to training restricted boltzmann machines.
\newblock In \emph{Neural networks: Tricks of the trade} (Springer). 599--619
\bibAnnoteFile{hinton2012practical}

\bibitem[{Hinton et~al.(1995)Hinton, Dayan, Frey, and Neal}]{hinton1995wake}
Hinton, G.~E., Dayan, P., Frey, B.~J., and Neal, R.~M. (1995).
\newblock The" wake-sleep" algorithm for unsupervised neural networks.
\newblock \emph{Science} 268, 1158--1161
\bibAnnoteFile{hinton1995wake}

\bibitem[{Hinton et~al.(1984)Hinton, Sejnowski, and
  Ackley}]{hinton1984boltzmann}
Hinton, G.~E., Sejnowski, T.~J., and Ackley, D.~H. (1984).
\newblock \emph{Boltzmann machines: Constraint satisfaction networks that
  learn} (Carnegie-Mellon University, Department of Computer Science
  Pittsburgh, PA)
\bibAnnoteFile{hinton1984boltzmann}

\bibitem[{Indiveri et~al.(2006)Indiveri, Chicca, and
  Douglas}]{indiveri2006vlsi}
Indiveri, G., Chicca, E., and Douglas, R.~J. (2006).
\newblock A vlsi array of low-power spiking neurons and bistable synapses with
  spike-timing dependent plasticity.
\newblock \emph{IEEE transactions on neural networks} 17
\bibAnnoteFile{indiveri2006vlsi}

\bibitem[{Indiveri et~al.(2011)Indiveri, Linares-Barranco, Hamilton,
  Van~Schaik, Etienne-Cummings, Delbruck et~al.}]{indiveri2011neuromorphic}
Indiveri, G., Linares-Barranco, B., Hamilton, T.~J., Van~Schaik, A.,
  Etienne-Cummings, R., Delbruck, T., et~al. (2011).
\newblock Neuromorphic silicon neuron circuits.
\newblock \emph{Frontiers in neuroscience} 5, 73
\bibAnnoteFile{indiveri2011neuromorphic}

\bibitem[{Jo et~al.(2010)Jo, Chang, Ebong, Bhadviya, Mazumder, and
  Lu}]{jo2010nanoscale}
Jo, S.~H., Chang, T., Ebong, I., Bhadviya, B.~B., Mazumder, P., and Lu, W.
  (2010).
\newblock Nanoscale memristor device as synapse in neuromorphic systems.
\newblock \emph{Nano letters} 10, 1297--1301
\bibAnnoteFile{jo2010nanoscale}

\bibitem[{Jones et~al.(2001--)Jones, Oliphant, Peterson
  et~al.}]{jones2014scipy}
[Dataset] Jones, E., Oliphant, T., Peterson, P., et~al. (2001--).
\newblock {SciPy}: Open source scientific tools for {Python}
\bibAnnoteFile{jones2014scipy}

\bibitem[{Jonke et~al.(2016)Jonke, Habenschuss, and Maass}]{jonke2016solving}
Jonke, Z., Habenschuss, S., and Maass, W. (2016).
\newblock Solving constraint satisfaction problems with networks of spiking
  neurons.
\newblock \emph{Frontiers in neuroscience} 10, 118
\bibAnnoteFile{jonke2016solving}

\bibitem[{Jordan et~al.(2017)Jordan, Petrovici, Breitwieser, Schemmel, Meier,
  Diesmann et~al.}]{jordan2017stochastic}
Jordan, J., Petrovici, M.~A., Breitwieser, O., Schemmel, J., Meier, K.,
  Diesmann, M., et~al. (2017).
\newblock Stochastic neural computation without noise.
\newblock \emph{arXiv preprint arXiv:1710.04931}
\bibAnnoteFile{jordan2017stochastic}

\bibitem[{Kullback and Leibler(1951)}]{kullback1951information}
Kullback, S. and Leibler, R.~A. (1951).
\newblock On information and sufficiency.
\newblock \emph{The annals of mathematical statistics} 22, 79--86
\bibAnnoteFile{kullback1951information}

\bibitem[{Kutschireiter et~al.(2017)Kutschireiter, Surace, Sprekeler, and
  Pfister}]{kutschireiter2017nonlinear}
Kutschireiter, A., Surace, S.~C., Sprekeler, H., and Pfister, J.-P. (2017).
\newblock Nonlinear bayesian filtering and learning: a neuronal dynamics for
  perception.
\newblock \emph{Scientific reports} 7, 8722
\bibAnnoteFile{kutschireiter2017nonlinear}

\bibitem[{Lande et~al.(1996)Lande, Ranjbar, Ismail, and Berg}]{lande1996analog}
Lande, T.~S., Ranjbar, H., Ismail, M., and Berg, Y. (1996).
\newblock An analog floating-gate memory in a standard digital technology.
\newblock In \emph{Proceedings of fifth international conference on
  microelectronics for neural networks} (IEEE), 271--276
\bibAnnoteFile{lande1996analog}

\bibitem[{LeCun et~al.(1998)LeCun, Bottou, Bengio, and
  Haffner}]{lecun1998gradient}
LeCun, Y., Bottou, L., Bengio, Y., and Haffner, P. (1998).
\newblock Gradient-based learning applied to document recognition.
\newblock \emph{Proceedings of the IEEE} 86, 2278--2324
\bibAnnoteFile{lecun1998gradient}

\bibitem[{Leng et~al.(2018)Leng, Martel, Breitwieser, Bytschok, Senn, Schemmel
  et~al.}]{leng2018spiking}
Leng, L., Martel, R., Breitwieser, O., Bytschok, I., Senn, W., Schemmel, J.,
  et~al. (2018).
\newblock Spiking neurons with short-term synaptic plasticity form superior
  generative networks.
\newblock \emph{Scientific reports} 8, 10651
\bibAnnoteFile{leng2018spiking}

\bibitem[{Loock(2006)}]{loock2006evaluierung}
Loock, J.-P. (2006).
\newblock Evaluierung eines floating gate analogspeichers f{\"u}r neuronale
  netze in single-poly umc 180nm cmos-prozess.
\newblock \emph{Diploma thesis (in German), University of Heidelberg,
  HD-KIP-06-47}
\bibAnnoteFile{loock2006evaluierung}

\bibitem[{Maaten and Hinton(2008)}]{maaten2008visualizing}
Maaten, L. v.~d. and Hinton, G. (2008).
\newblock Visualizing data using t-sne.
\newblock \emph{Journal of machine learning research} 9, 2579--2605
\bibAnnoteFile{maaten2008visualizing}

\bibitem[{Mainen and Sejnowski(1995)}]{mainen1995reliability}
Mainen, Z.~F. and Sejnowski, T.~J. (1995).
\newblock Reliability of spike timing in neocortical neurons.
\newblock \emph{Science} 268, 1503--1506
\bibAnnoteFile{mainen1995reliability}

\bibitem[{Mead(1990)}]{mead1990neuromorphic}
Mead, C. (1990).
\newblock Neuromorphic electronic systems.
\newblock \emph{Proceedings of the IEEE} 78, 1629--1636
\bibAnnoteFile{mead1990neuromorphic}

\bibitem[{Merolla et~al.(2014)Merolla, Arthur, Alvarez-Icaza, Cassidy, Sawada,
  Akopyan et~al.}]{merolla2014million}
Merolla, P.~A., Arthur, J.~V., Alvarez-Icaza, R., Cassidy, A.~S., Sawada, J.,
  Akopyan, F., et~al. (2014).
\newblock A million spiking-neuron integrated circuit with a scalable
  communication network and interface.
\newblock \emph{Science} 345, 668--673
\bibAnnoteFile{merolla2014million}

\bibitem[{Millner(2012)}]{millner2012development}
Millner, S. (2012).
\newblock \emph{Development of a multi-compartment neuron model emulation}.
\newblock Ph.D. thesis
\bibAnnoteFile{millner2012development}

\bibitem[{Millner et~al.(2010)Millner, Gr\"{u}bl, Meier, Schemmel, and
  Schwartz}]{millner2010AdEx}
Millner, S., Gr\"{u}bl, A., Meier, K., Schemmel, J., and Schwartz, M.-O.
  (2010).
\newblock A {VLSI} implementation of the adaptive exponential
  integrate-and-fire neuron model.
\newblock In \emph{Adv Neur In}, eds. J.~Lafferty, C.~K.~I. Williams,
  J.~Shawe-Taylor, R.~Zemel, and A.~Culotta. vol.~23, 1642--1650
\bibAnnoteFile{millner2010AdEx}

\bibitem[{Orb{\'a}n et~al.(2016)Orb{\'a}n, Berkes, Fiser, and
  Lengyel}]{orban2016neural}
Orb{\'a}n, G., Berkes, P., Fiser, J., and Lengyel, M. (2016).
\newblock Neural variability and sampling-based probabilistic representations
  in the visual cortex.
\newblock \emph{Neuron} 92, 530--543
\bibAnnoteFile{orban2016neural}

\bibitem[{Pedroni et~al.(2016)Pedroni, Das, Arthur, Merolla, Jackson, Modha
  et~al.}]{pedroni2016mapping}
Pedroni, B.~U., Das, S., Arthur, J.~V., Merolla, P.~A., Jackson, B.~L., Modha,
  D.~S., et~al. (2016).
\newblock Mapping generative models onto a network of digital spiking neurons.
\newblock \emph{IEEE transactions on biomedical circuits and systems} 10,
  837--854
\bibAnnoteFile{pedroni2016mapping}

\bibitem[{Petrovici(2016)}]{petrovici2016form}
Petrovici, M.~A. (2016).
\newblock \emph{Form Versus Function: Theory and Models for Neuronal
  Substrates} (Springer)
\bibAnnoteFile{petrovici2016form}

\bibitem[{Petrovici et~al.(2013)Petrovici, Bill, Bytschok, Schemmel, and
  Meier}]{petrovici2013stochastic}
Petrovici, M.~A., Bill, J., Bytschok, I., Schemmel, J., and Meier, K. (2013).
\newblock Stochastic inference with deterministic spiking neurons.
\newblock \emph{arXiv preprint arXiv:1311.3211}
\bibAnnoteFile{petrovici2013stochastic}

\bibitem[{Petrovici et~al.(2016)Petrovici, Bill, Bytschok, Schemmel, and
  Meier}]{petrovici2016stochastic}
Petrovici, M.~A., Bill, J., Bytschok, I., Schemmel, J., and Meier, K. (2016).
\newblock Stochastic inference with spiking neurons in the high-conductance
  state.
\newblock \emph{Physical Review E} 94, 042312
\bibAnnoteFile{petrovici2016stochastic}

\bibitem[{Petrovici et~al.(2017{\natexlab{a}})Petrovici, Schmitt, Kl{\"a}hn,
  St{\"o}ckel, Schroeder, Bellec et~al.}]{petrovici2017pattern}
Petrovici, M.~A., Schmitt, S., Kl{\"a}hn, J., St{\"o}ckel, D., Schroeder, A.,
  Bellec, G., et~al. (2017{\natexlab{a}}).
\newblock Pattern representation and recognition with accelerated analog
  neuromorphic systems.
\newblock In \emph{2017 IEEE International Symposium on Circuits and Systems
  (ISCAS)} (IEEE), 1--4
\bibAnnoteFile{petrovici2017pattern}

\bibitem[{Petrovici et~al.(2017{\natexlab{b}})Petrovici, Schroeder,
  Breitwieser, Gr{\"u}bl, Schemmel, and Meier}]{petrovici2017robustness}
Petrovici, M.~A., Schroeder, A., Breitwieser, O., Gr{\"u}bl, A., Schemmel, J.,
  and Meier, K. (2017{\natexlab{b}}).
\newblock Robustness from structure: Inference with hierarchical spiking
  networks on analog neuromorphic hardware.
\newblock In \emph{Neural Networks (IJCNN), 2017 International Joint Conference
  on} (IEEE), 2209--2216
\bibAnnoteFile{petrovici2017robustness}

\bibitem[{Petrovici et~al.(2015)Petrovici, St{\"o}ckel, Bytschok, Bill, Pfeil,
  Schemmel et~al.}]{petrovici2015sampling}
Petrovici, M.~A., St{\"o}ckel, D., Bytschok, I., Bill, J., Pfeil, T., Schemmel,
  J., et~al. (2015).
\newblock Fast sampling with neuromorphic hardware.
\newblock In \emph{Advances in Neural Information Processing Systems (NIPS)}.
  vol.~28
\bibAnnoteFile{petrovici2015sampling}

\bibitem[{Petrovici et~al.(2014)Petrovici, Vogginger, M{\"u}ller, Breitwieser,
  Lundqvist, Muller et~al.}]{petrovici2014characterization}
Petrovici, M.~A., Vogginger, B., M{\"u}ller, P., Breitwieser, O., Lundqvist,
  M., Muller, L., et~al. (2014).
\newblock Characterization and compensation of network-level anomalies in
  mixed-signal neuromorphic modeling platforms.
\newblock \emph{PloS one} 9, e108590
\bibAnnoteFile{petrovici2014characterization}

\bibitem[{Pfeil et~al.(2013)Pfeil, Gr{\"u}bl, Jeltsch, M{\"u}ller, M{\"u}ller,
  Petrovici et~al.}]{pfeil2013six}
Pfeil, T., Gr{\"u}bl, A., Jeltsch, S., M{\"u}ller, E., M{\"u}ller, P.,
  Petrovici, M.~A., et~al. (2013).
\newblock Six networks on a universal neuromorphic computing substrate.
\newblock \emph{Frontiers in neuroscience} 7, 11
\bibAnnoteFile{pfeil2013six}

\bibitem[{Pfeil et~al.(2016)Pfeil, Jordan, Tetzlaff, Gr{\"u}bl, Schemmel,
  Diesmann et~al.}]{pfeil2016effect}
Pfeil, T., Jordan, J., Tetzlaff, T., Gr{\"u}bl, A., Schemmel, J., Diesmann, M.,
  et~al. (2016).
\newblock Effect of heterogeneity on decorrelation mechanisms in spiking neural
  networks: A neuromorphic-hardware study.
\newblock \emph{Physical Review X} 6, 021023
\bibAnnoteFile{pfeil2016effect}

\bibitem[{Pouget et~al.(2013)Pouget, Beck, Ma, and
  Latham}]{pouget2013probabilistic}
Pouget, A., Beck, J.~M., Ma, W.~J., and Latham, P.~E. (2013).
\newblock Probabilistic brains: knowns and unknowns.
\newblock \emph{Nature neuroscience} 16, 1170
\bibAnnoteFile{pouget2013probabilistic}

\bibitem[{Probst et~al.(2015)Probst, Petrovici, Bytschok, Bill, Pecevski,
  Schemmel et~al.}]{probst2015probabilistic}
Probst, D., Petrovici, M.~A., Bytschok, I., Bill, J., Pecevski, D., Schemmel,
  J., et~al. (2015).
\newblock Probabilistic inference in discrete spaces can be implemented into
  networks of lif neurons.
\newblock \emph{Frontiers in computational neuroscience} 9, 13
\bibAnnoteFile{probst2015probabilistic}

\bibitem[{Qiao et~al.(2015)Qiao, Mostafa, Corradi, Osswald, Stefanini,
  Sumislawska et~al.}]{qiao2015reconfigurable}
Qiao, N., Mostafa, H., Corradi, F., Osswald, M., Stefanini, F., Sumislawska,
  D., et~al. (2015).
\newblock A reconfigurable on-line learning spiking neuromorphic processor
  comprising 256 neurons and 128k synapses.
\newblock \emph{Frontiers in neuroscience} 9, 141
\bibAnnoteFile{qiao2015reconfigurable}

\bibitem[{Reinagel and Reid(2002)}]{reinagel2002precise}
Reinagel, P. and Reid, R.~C. (2002).
\newblock Precise firing events are conserved across neurons.
\newblock \emph{Journal of Neuroscience} 22, 6837--6841
\bibAnnoteFile{reinagel2002precise}

\bibitem[{Salakhutdinov(2010)}]{salakhutdinov2010learning}
Salakhutdinov, R. (2010).
\newblock Learning deep boltzmann machines using adaptive mcmc.
\newblock In \emph{Proceedings of the 27th International Conference on Machine
  Learning (ICML-10)}. 943--950
\bibAnnoteFile{salakhutdinov2010learning}

\bibitem[{Savitzky and Golay(1964)}]{savitzky1964smoothing}
Savitzky, A. and Golay, M.~J. (1964).
\newblock Smoothing and differentiation of data by simplified least squares
  procedures.
\newblock \emph{Analytical chemistry} 36, 1627--1639
\bibAnnoteFile{savitzky1964smoothing}

\bibitem[{Schemmel et~al.(2010)Schemmel, Br{\"u}derle, Gr{\"u}bl, Hock, Meier,
  and Millner}]{schemmel2010wafer}
Schemmel, J., Br{\"u}derle, D., Gr{\"u}bl, A., Hock, M., Meier, K., and
  Millner, S. (2010).
\newblock A wafer-scale neuromorphic hardware system for large-scale neural
  modeling.
\newblock In \emph{Circuits and systems (ISCAS), proceedings of 2010 IEEE
  international symposium on} (IEEE), 1947--1950
\bibAnnoteFile{schemmel2010wafer}

\bibitem[{Schmitt et~al.(2017)Schmitt, Kl{\"a}hn, Bellec, Gr{\"u}bl, Guettler,
  Hartel et~al.}]{schmitt2017neuromorphic}
Schmitt, S., Kl{\"a}hn, J., Bellec, G., Gr{\"u}bl, A., Guettler, M., Hartel,
  A., et~al. (2017).
\newblock Neuromorphic hardware in the loop: Training a deep spiking network on
  the brainscales wafer-scale system.
\newblock In \emph{Neural Networks (IJCNN), 2017 International Joint Conference
  on} (IEEE), 2227--2234
\bibAnnoteFile{schmitt2017neuromorphic}

\bibitem[{Schmuker et~al.(2014)Schmuker, Pfeil, and
  Nawrot}]{schmuker2014neuromorphic}
Schmuker, M., Pfeil, T., and Nawrot, M.~P. (2014).
\newblock A neuromorphic network for generic multivariate data classification.
\newblock \emph{Proceedings of the National Academy of Sciences} 111,
  2081--2086
\bibAnnoteFile{schmuker2014neuromorphic}

\bibitem[{Sutskever and Hinton(2007)}]{sutskever2007learning}
Sutskever, I. and Hinton, G. (2007).
\newblock Learning multilevel distributed representations for high-dimensional
  sequences.
\newblock In \emph{Artificial intelligence and statistics}. 548--555
\bibAnnoteFile{sutskever2007learning}

\bibitem[{Taylor and Hinton(2009)}]{taylor2009factored}
Taylor, G.~W. and Hinton, G.~E. (2009).
\newblock Factored conditional restricted boltzmann machines for modeling
  motion style.
\newblock In \emph{Proceedings of the 26th annual international conference on
  machine learning} (ACM), 1025--1032
\bibAnnoteFile{taylor2009factored}

\bibitem[{Tetzlaff et~al.(2012)Tetzlaff, Helias, Einevoll, and
  Diesmann}]{tetzlaff2012decorrelation}
Tetzlaff, T., Helias, M., Einevoll, G.~T., and Diesmann, M. (2012).
\newblock Decorrelation of neural-network activity by inhibitory feedback.
\newblock \emph{PLoS computational biology} 8, e1002596
\bibAnnoteFile{tetzlaff2012decorrelation}

\bibitem[{Toups et~al.(2012)Toups, Fellous, Thomas, Sejnowski, and
  Tiesinga}]{toups2012multiple}
Toups, J.~V., Fellous, J.-M., Thomas, P.~J., Sejnowski, T.~J., and Tiesinga,
  P.~H. (2012).
\newblock Multiple spike time patterns occur at bifurcation points of membrane
  potential dynamics.
\newblock \emph{PLoS computational biology} 8, e1002615
\bibAnnoteFile{toups2012multiple}

\bibitem[{Waldrop(2016)}]{waldrop2016chips}
Waldrop, M.~M. (2016).
\newblock The chips are down for {M}oore's law.
\newblock \emph{Nature News} 530, 144
\bibAnnoteFile{waldrop2016chips}

\bibitem[{Wunderlich et~al.(2019)Wunderlich, Kungl, M{\"u}ller, Hartel,
  Stradmann, Aamir et~al.}]{wunderlich2018demonstrating}
Wunderlich, T., Kungl, A.~F., M{\"u}ller, E., Hartel, A., Stradmann, Y., Aamir,
  S.~A., et~al. (2019).
\newblock Demonstrating advantages of neuromorphic computation: A pilot study.
\newblock \emph{Frontiers in Neuroscience} 13, 260
\bibAnnoteFile{wunderlich2018demonstrating}

\bibitem[{Xiao et~al.(2017)Xiao, Rasul, and Vollgraf}]{xiao2017online}
[Dataset] Xiao, H., Rasul, K., and Vollgraf, R. (2017).
\newblock Fashion-mnist: a novel image dataset for benchmarking machine
  learning algorithms
\bibAnnoteFile{xiao2017online}

\bibitem[{Zoschke et~al.(2017)Zoschke, G\"{u}ttler, B\"{o}ttcher, Gr\"{u}bl,
  Husmann, Schemmel et~al.}]{zoschkeguettler2017rdlembedding}
Zoschke, K., G\"{u}ttler, M., B\"{o}ttcher, L., Gr\"{u}bl, A., Husmann, D.,
  Schemmel, J., et~al. (2017).
\newblock Full wafer redistribution and wafer embedding as key technologies for
  a multi-scale neuromorphic hardware cluster.
\newblock \emph{EPTC 2017}
\bibAnnoteFile{zoschkeguettler2017rdlembedding}

\end{thebibliography}

\raggedbottom

\section*{Tables}
\FloatBarrier

\begin{table}
    \caption{
        \textbf{Description of the neuron and synapse model.}
        The variables are described including their numerical values in the experiment in \cref{table:neuron}.
            }
      \begin{center}
        \begin{tabular}{llr}
            \toprule
            Type & \multicolumn{2}{l}{Leaky integrate-and-fire (LIF), conductance based synapse, exponential kernel} \\
            \midrule
            Subthreshold dynamics & \multicolumn{2}{l}{Subthreshold dynamics $[t\notin[t_\mathrm{sp}, t_\mathrm{sp} + \tau_\mathrm{ref})]:$} \\
            & \multicolumn{2}{l}{$C_\mathrm{m}(\mathrm{d}/\mathrm{d}t)u(t)=-g_\mathrm{l}[u(t) - E_\mathrm{leak}] - g_\mathrm{syn}^\mathrm{inh}(t)[u(t) - E_\mathrm{inh}] - g_\mathrm{syn}^\mathrm{exc}(t)[u(t) - E_\mathrm{exc}]$} \\
            Reset and refractoriness & \multicolumn{2}{l}{$[t\in[t_\mathrm{sp}, t_\mathrm{sp} + \tau_\mathrm{ref})]:$} \\
            & \multicolumn{2}{l}{$u(t) = V_\mathrm{reset}$} \\
            & \multicolumn{2}{l}{This model was emulated on the BrainScaleS system~\citep{schemmel2010wafer} } \\
            Spiking & \multicolumn{2}{l}{ If $u(t)$ crosses $V_\mathrm{thresh}$ from below at $t=t_\mathrm{sp}$,} \\
            & neuron emits a spike with timestamp $t_\mathrm{sp}$ \\
            Synapse dynamics & \multicolumn{2}{l}{For each presynaptic spike at $t_\mathrm{sp}:$} \\
            & \multicolumn{2}{l}{$g_\mathrm{syn}(t) = J \exp[-(t-t_\mathrm{sp} - d)/(\tau_\mathrm{syn})]\theta(t-t_\mathrm{sp}-d)$} \\
            & \multicolumn{2}{l}{where $J$ is the synaptic weight, $d$ the synaptic delay and $\theta$ the Heaviside function}\\
            & \multicolumn{2}{l}{This model was emulated on the BrainScaleS system~\citep{schemmel2010wafer}}\\
            \bottomrule

        \end{tabular}
        \end{center}
    \label{table:network}
\end{table}

\begin{table}
    \caption{
        \textbf{Neuron parameters.}
        Parameters of the network setup specified in \cref{table:network}.
        The analog parameters are shown as specified in the software setup and not as realized on the hardware.
        For details on the calibration procedure see, e.g., \citep{schmitt2017neuromorphic}.
        \emph{Legend:} $^{*}$ the calibration of the membrane time constant was not available at the time of this work, and the corresponding technical parameter was set to the smallest available value instead (fastest possible membrane dynamics for each neuron).
        }
    \begin{center}
        \begin{tabular}{lll}
            \toprule
            \textbf{A} & \multicolumn{2}{c}{\textbf{Sampling neuron}} \\
            \midrule
            \textit{Name} & \textit{Value} & \textit{Description} \\
            $V_\mathrm{reset}$ & \SI{-35}{mV} & reset potential\\
            $E_\mathrm{leak}$ & \SI{-20}{mV} & resting potential\\
            $V_\mathrm{thresh}$ & \SI{-20}{mV} & threshold potential\\
            $E_\mathrm{inh}$ & \SI{-100}{mV} & inhibitory reversal potential\\
            $E_\mathrm{exc}$ & \SI{60}{mV} & excitatory reversal potential\\
            $\tau_\mathrm{ref}$ & \SI{4}{ms} & refractory time\\
            $\tau_\mathrm{mem}$ & ca. \SI{7}{ms} & membrane time constant$^{*}$\\
            $C_\mathrm{mem}$ & \SI{0.2}{nF} & membrane capacity\\
            $\tau_\mathrm{syn}^\mathrm{exc}$ & \SI{8}{ms} & excitatory synaptic time constant \\
            $\tau_\mathrm{syn}^\mathrm{inh}$ & \SI{8}{ms} & inhibitory synaptic time constant \\
            \midrule
            \textbf{B} & \multicolumn{2}{c}{\textbf{Bias neuron}} \\
            \midrule
            \textit{Name} & \textit{Value} & \textit{Description} \\
            $V_\mathrm{reset}$ & \SI{-30}{mV} & reset potential\\
            $E_\mathrm{leak}$ & \SI{60}{mV} & resting potential\\
            $V_\mathrm{thresh}$ & \SI{-20}{mV} & threshold potential\\
            $E_\mathrm{inh}$ & \SI{-100}{mV} & inhibitory reversal potential\\
            $E_\mathrm{exc}$ & \SI{60}{mV} & excitatory reversal potential\\
            $\tau_\mathrm{ref}$ & \SI{1.5}{ms} & refractory time\\
            $\tau_\mathrm{mem}$ & ca. \SI{7}{ms} & membrane time constant$^{*}$\\
            $C_\mathrm{mem}$ & \SI{0.2}{nF} & membrane capacity\\
            $\tau_\mathrm{syn}^\mathrm{exc}$ & \SI{5}{ms}& excitatory synaptic time constant \\
            $\tau_\mathrm{syn}^\mathrm{inh}$ & \SI{5}{ms}& inhibitory synaptic time constant \\
            \midrule
            \textbf{C} & \multicolumn{2}{c}{\textbf{Neurons of the random network}} \\
            \midrule
            \textit{Name} & \textit{Value} & \textit{Description (all analog)} \\
            $V_\mathrm{reset}$ & \SI{-60}{mV} & reset potential\\
            $E_\mathrm{leak}$ & \SI{-10}{mV} & resting potential\\
            $V_\mathrm{thresh}$ & \SI{-20}{mV} & threshold potential\\
            $E_\mathrm{inh}$ & \SI{-100}{mV} & inhibitory reversal potential\\
            $E_\mathrm{exc}$ & \SI{60}{mV} & excitatory reversal potential\\
            $\tau_\mathrm{ref}$ & \SI{4}{ms} & refractory time\\
            $\tau_\mathrm{mem}$ & ca. \SI{7}{ms} & membrane time constant$^{*}$\\
            $C_\mathrm{mem}$ & \SI{0.2}{nF} & membrane capacity\\
            $\tau_\mathrm{syn}^\mathrm{exc}$ & \SI{8}{ms}& excitatory synaptic time constant \\
            $\tau_\mathrm{syn}^\mathrm{inh}$ & \SI{8}{ms}& inhibitory synaptic time constant \\
            \midrule
            \textbf{D} & \multicolumn{2}{c}{\textbf{Synapse}} \\
            \midrule
            \textit{Name} & \textit{Value} & \textit{Description} \\
            $w_\mathrm{bias}$ & [0,15] & synaptic bias weight in hardware values (digital) \\
            $w_\mathrm{network}$ & [0,15] & synaptic network weight in hardware values (digital) \\
            $d$ & on the order of \SI{1}{ms} (uncalibrated) & synaptic delay, estimated in~\citep{schemmel2010wafer} \\
            \bottomrule
        \end{tabular}
    \end{center}
    \label{table:neuron}
\end{table}

\begin{table}
    \caption{
        \textbf{Parameters for learning.}
        We did not carry any systematic hyper-parameter optimization.
        Note that the used learning parameters in the experiments in \cref{sec:distr} are not directly comparable because the different statistics of the background noise (Poisson or random network) correspond to different effective learning rates.
        }
      \begin{center}
      \begin{tabular}{lllll}
            \toprule
            \textit{Experiment} & \textit{Learning rate} & \textit{Momentum factor} & \textit{minibatch-size}  & \textit{Initial $(\mathbf{W},\mathbf{b})$} \\
            \midrule
            target distribution, Poisson & 1.0 & 0.6 & - & $\mathcal{U}(-15,15)$ \\
            target distribution, random network & 0.5 & 0.6 & - & $\mathcal{U}(-15,15)$ \\
            rMNIST & 0.4 & 0.6 & 7/class & pre-trained \\
            rFMNIST & 0.4 & 0.6 & 7/class & pre-trained \\
            \bottomrule
        \end{tabular}
        \end{center}
    \label{table:learnParam}
\end{table}

\begin{table}
    \caption{
        \textbf{Network parameters.}
        Parameters are shown for the three different cases described in the manuscript:
        \textbf{A} Target Boltzmann distribution, Poisson noise.
        \textbf{B} Target Boltzmann distribution, random network for stochasticity.
        \textbf{C} Learning from data, random network for stochasticity.
        { Note that the \textit{in-degree}, sometimes also referred to as a \textit{fan-in factor}, represents a neuron's number of pre-synaptic partners coming from some specific population.}}
      \begin{center}
      \begin{tabular}{lll}
            \toprule
            \textbf{A} & \multicolumn{2}{c}{\textbf{Probability distribution with Poisson Noise}} \\
            \midrule
            \textit{Name} & \textit{Value} & \textit{Description} \\
            $N_\mathrm{s}$ & 5 & number of sampling neurons \\
            $N_\mathrm{b}$ & 1 & number of bias neurons \\
            $N_\mathrm{r}$ & 0 & number of random neurons \\
            $K_\mathrm{RN}$ & - & within-population in-degree of neurons in the random network \\
            $K_\mathrm{noise}$ & - & in-degree of sampling neurons from the random network \\
            $w_\mathrm{RN}$ & - & synaptic weights in the random network \\
            & & in hardware units \\
            $\nu^{\mathrm{e/i}}_\mathrm{Poisson}$ & \SI{300}{Hz} & Poisson frequency to sampling neurons per synapse type \\
            \midrule
            \textbf{B} & \multicolumn{2}{c}{\textbf{Probability distribution with random network}} \\
            \midrule
            \textit{Name} & \textit{Value} & \textit{Description} \\
            $N_\mathrm{s}$ & 5 & number of sampling neurons \\
            $N_\mathrm{b}$ & 1 & number of bias neurons \\
            $N_\mathrm{r}$ & 200 & number of random neurons \\
            $K_\mathrm{RN}$ & 20 & within-population in-degree of neurons in the random network \\
            $K_\mathrm{noise}$ & 15 & in-degree of sampling neurons from the random network \\
            $w_\mathrm{RN}$ & 10 & synaptic weights in the random network \\
            & & in hardware units \\
            $\nu^{\mathrm{e/i}}_\mathrm{Poisson}$ & - & Poisson frequency to sampling neurons per synapse type \\
            \midrule
            \textbf{C} & \multicolumn{2}{c}{\textbf{High-dimensional dataset}} \\
            \midrule
            \textit{Name} & \textit{Value} & \textit{Description} \\
            $N_\mathrm{s}$ & $\{ 207,208 \}$ & number of sampling neurons, \{ rFMNIST, rMNIST \} \\
            $N_\mathrm{b}$ & 1 & number of bias neurons \\
            $N_\mathrm{r}$ & 400 & number of random neurons \\
            $K_\mathrm{RN}$ & 20 & within-population in-degree of neurons in the random network \\
            $K_\mathrm{noise}$ & 15 & in-degree of sampling neurons from the random network \\
            $w_\mathrm{RN}$ & 10 & synaptic weights in the random network \\
            & & in hardware units \\
            $\nu^{\mathrm{e/i}}_\mathrm{Poisson}$ & - & Poisson frequency to sampling neurons per synapse type \\
            \bottomrule
    \end{tabular}
    \end{center}
    \label{table:netParam}
\end{table}

\end{document}